\documentclass{article}
\usepackage{ijcai22}

\usepackage{times}
\usepackage{soul}
\usepackage{url}
\usepackage[hidelinks]{hyperref}
\usepackage[utf8]{inputenc}
\usepackage[small]{caption}
\usepackage{graphicx}
\usepackage{amsmath}
\usepackage{amsthm}
\usepackage{amssymb}
\usepackage{booktabs}
\usepackage{algorithm}
\usepackage{algorithmic}
\usepackage{multirow}
\usepackage{xcolor}
\usepackage{enumitem}
\title{Leveraging class abstraction for commonsense reinforcement learning via residual policy gradient methods}

\author{
Niklas H\"opner$^1$\footnote{Contact Author}\and
Ilaria Tiddi $^2$\And
Herke van Hoof $^1$ \\
\affiliations
$^1$University of Amsterdam\\
$^2$Vrije Universiteit Amsterdam\\
\emails
\{n.r.hopner, h.c.vanhoof\}@uva.nl,
i.tiddi@vu.nl,
}

\begin{document}

\maketitle

\begin{abstract}

Enabling reinforcement learning (RL) agents to leverage a knowledge base while learning from experience promises to advance RL in knowledge intensive domains. However, it has proven difficult to leverage knowledge that is not manually tailored to the environment. We propose to use the subclass relationships present in open-source knowledge graphs to abstract away from specific objects. We develop a residual policy gradient method that is able to integrate knowledge across different abstraction levels in the class hierarchy. Our method results in improved sample efficiency and generalisation to unseen objects in commonsense games, but we also investigate failure modes, such as excessive noise in the extracted class knowledge or environments with little class structure\footnote{Code can be found at https://github.com/NikeHop/CSRL}.

\end{abstract}

\section{Introduction}


Deep reinforcement learning (DRL) has enabled us to optimise control policies in MDPs with high-dimensional state and action spaces such as in game-play \cite{silver_go} and robotics \cite{continuous_control}. Two main hindrances in bringing deep reinforcement learning to the real world are the sample inefficiency and poor generalisation performance of current methods \cite{survey_generalisation}. Amongst other approaches, including prior knowledge in the learning process of the agent promises to alleviate these obstacles and move reinforcement learning (RL) from a tabula rasa method to more human-like learning. Depending on the area of research, prior knowledge representations can vary from pretrained embeddings or weights \cite{BERT} to symbolic knowledge representations such as logics \cite{logic} and knowledge graphs (KGs) \cite{story_completion}. While the former are easier to integrate into deep neural network-based algorithms, they lack specificity, abstractness, robustness and interpretability \cite{boxology}. 

One type of prior knowledge that is hard to obtain for purely data-driven methods is commonsense knowledge. Equipping reinforcement learning agents with commonsense or world knowledge is an important step towards improved human-machine interaction \cite{HI}, as interesting interactions demand machines to access prior knowledge not learnable from experience. Commonsense games \cite{wordcraft,twc} have emerged as a testbed for methods that aim at integrating commonsense knowledge into a RL agent. Prior work has focused on augmenting the state by extracting subparts of ConceptNet \cite{twc}. Performance only improved when the extracted knowledge was tailored to the environment. Here, we focus on knowledge that is automatically extracted and should be useful across a range of commonsense games.

Humans abstract away from specific objects using classes, which allows them to learn behaviour at class-level and generalise to unseen objects \cite{yee2019abstraction}. Since commonsense games deal with real-world entities, we look at the problem of leveraging subclass knowledge from open-source KGs to improve sample efficiency and generalisation of an agent in commonsense games. We use subclass knowledge to formulate a state abstraction, that aggregates states depending on which classes are present in a given state. This state abstraction might not preserve all information necessary to act optimally in a state. Therefore, a method is needed that learns to integrate useful knowledge over a sequence of more and more fine-grained state representations. We show how a naive ensemble approach can fail to correctly integrate information from imperfectly abstracted states, and design a residual learning approach that is forced to learn the difference between policies over adjacent abstraction levels. The properties of both approaches are first studied in a toy setting where the effectiveness of class-based abstractions can be controlled. We then show that if a commonsense game is governed by class structure, the agent is more sample efficient and generalises better to unseen objects, outperforming embedding approaches and methods augmenting the state with subparts of ConceptNet. However, learning might be hampered, if the extracted class knowledge aggregates objects incorrectly. To summarise, our key contributions are:
\begin{itemize}[leftmargin=*]
    \item we use the subclass relationship from open-source KGs to formulate a state abstraction for commonsense games;
    \item we propose a residual learning approach that can be integrated with policy gradient algorithms to leverage imperfect state abstractions;
    \item we show that in environments with class structure our method leads to more sample efficient learning and better generalisation to unseen objects.
\end{itemize}

\section{Related Work}

We introduce the resources available to include commonsense knowledge and the attempts that have been made by prior work to leverage these resources. Since the class knowledge is extracted from knowledge graphs, work on including KGs in deep neural network based architectures is discussed. The setting considered here also offers a new perspective on state abstraction in reinforcement learning. 
 
\paragraph{Commonsense KGs.}
KGs store facts in form of entity-relation-entity triplets. Often a KG is constructed to capture either a general area of knowledge such as commonsense \cite{cskg}, or more domain specific knowledge like the medical domain \cite{depression}. While ConceptNet \cite{conceptnet} tries to represent all commonsense knowledge, others focus on specific parts such as cause and effect relations \cite{atomic}. Manually designed KGs \cite{wordnet} are less error prone, but provide less coverage and are more costly to design, making hybrid approaches popular \cite{conceptnet}. Here, we focus on WordNet \cite{wordnet}, ConceptNet and DBpedia \cite{dbpedia} and study how the quality of their class knowledge affects our method. Representing KGs in vector form can be achieved via knowledge graph embedding techniques \cite{poincare}, where embeddings can be trained from scratch or word embeddings are finetuned \cite{numberbatch}. Hyperbolic embeddings \cite{poincare} capture the hierarchical structure of WordNet given by the hypernym relation between two nouns and are investigated as an alternative method to include class prior knowledge.

\paragraph{Commonsense Games.}
To study the problem of integrating commonsense knowledge into an RL agent, commonsense games have recently been introduced \cite{wordcraft,twc}. Prior methods have focused on leveraging knowledge graph embeddings \cite{wordcraft} or augmenting the state representation via an extracted subpart of ConceptNet \cite{twc}. While knowledge graph embeddings improve performance more than GloVe word embeddings \cite{glove}, the knowledge graph they are based on is privileged game information. Extracting task-relevant knowledge from ConceptNet automatically is challenging. If the knowledge is manually specified, sample efficiency improves but heuristic extraction rules hamper learning. No method that learns to extract useful knowledge exists yet. The class knowledge we use here is not tailored to the environment, and therefore should hold across a range of environments.

\paragraph{Integrating KGs into deep learning architectures.} The problem of integrating knowledge present in a knowledge graph into a learning algorithm based on deep neural networks has mostly been studied by the natural language community \cite{commonsense_overview}. Use-cases include, but are not limited to, open-dialog \cite{Open_Dialog_1}, task-oriented dialogue \cite{task1} and story completion \cite{story_completion}. Most methods are based on an attention mechanism over parts of the knowledge base \cite{Open_Dialog_1,task1}. Two key differences are that the knowledge graphs used are curated for the task, and therefore contain little noise. Additionally, most tasks are framed as supervised learning problems where annotation about correct reasoning patterns are given. Here, we extract knowledge from open-source knowledge graphs and have to deal with errors in the class structure due to problems with entity reconciliation and incompleteness of knowledge. 

\paragraph{State abstraction in RL.} State abstraction aims to partition the state space of a base Markov Decision Process (MPD) into abstract states to reduce the complexity of the state space on which a policy is learnt \cite{li_state}. Different criteria for aggregating states have been proposed \cite{bisimulation}. They guarantee that an optimal policy learnt for the abstract MDP remains optimal for the base MDP. To leverage state abstraction an aggregation function has to be learned \cite{block_mdps}, which either needs additional samples or is performed on-policy leading to a potential collapse of the aggregation function \cite{neurips_bisimulation}. The case in which an approximate state abstraction is given as prior knowledge has not been looked at yet. The abstraction given here must not satisfy any consistency criteria and can consist of multiple abstraction levels. A method that is capable of integrating useful knowledge from each abstraction level is needed.




\section{Problem Setting} \label{sec:problem_formulation}
\begin{figure}
    \centering

    \begin{minipage}{0.6\linewidth}
        \includegraphics[width=\textwidth]{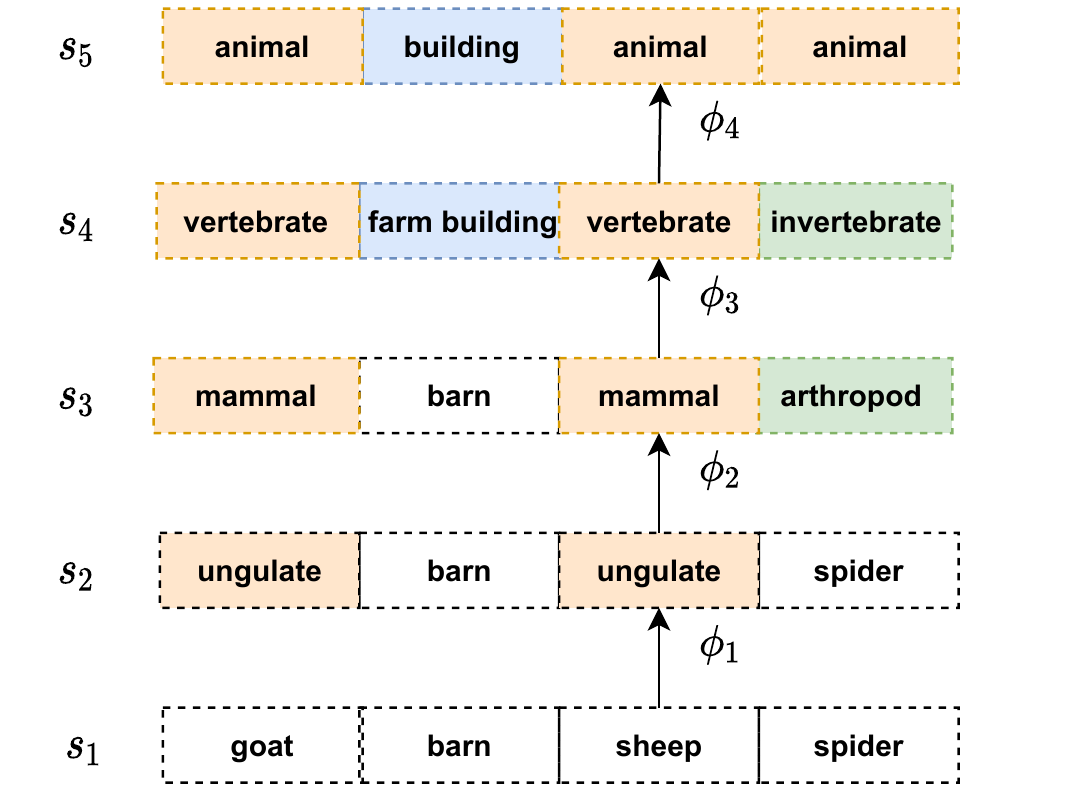}
        \caption*{}
    \end{minipage}
    \begin{minipage}{0.39\linewidth}
        \includegraphics[width=\textwidth]{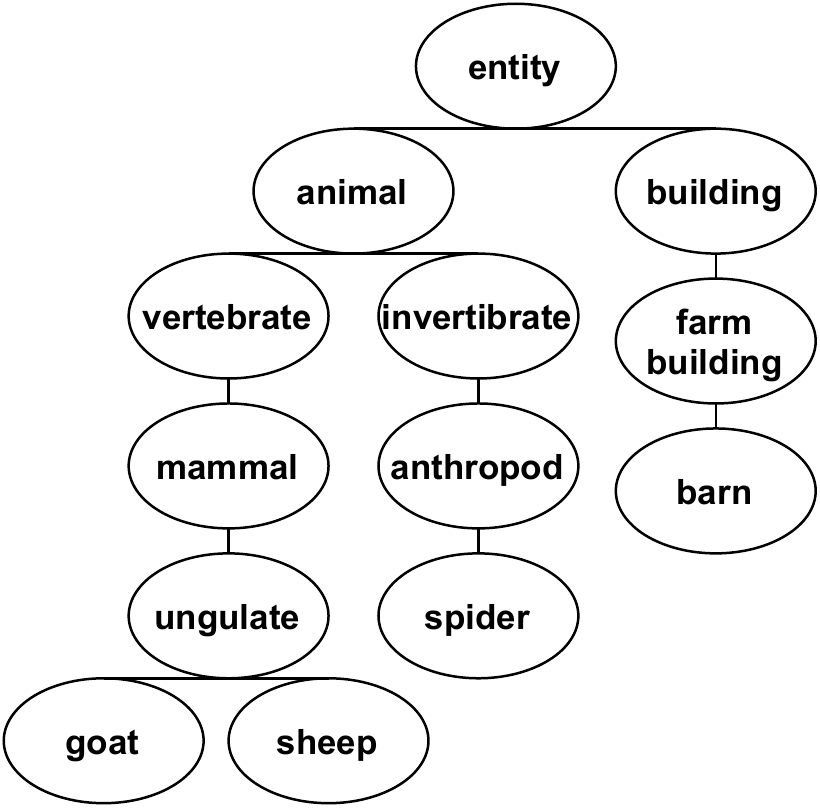}
        \caption*{}
    \end{minipage}
    \vspace{-0.75cm}
    \caption{(Left) Visualisation of a state and its abstractions in Wordcraft based on classes extracted from WordNet. Colours indicate that an object is mapped to a class, where the same colour refers to the same class within an abstraction. (Right) Subgraph of the class tree that is used to determine the state abstractions.}
    \label{fig:example_abstraction}
\end{figure}
Reinforcement learning enables us to learn optimal behaviour in an MDP $M=(S,A,R,T,\gamma)$ with state space $S$, action space $A$, reward function $R: S\times A \rightarrow \mathbb{R}$, discount factor $\gamma$ and transition function $T: S\times A \rightarrow \Delta{S}$ , where $\Delta{S}$ represents the set of probability distributions over the space $S$. The goal is to learn from experience a policy $\pi:S \rightarrow \Delta{A}$ that optimises the objective:
\begin{equation}
    J(\pi) = \mathbb{E}_{\pi}\left[\sum_{t=0}^{\infty}\gamma^{t} R_{t}\right] = \mathbb{E}_{\pi}[G],
\end{equation}
where $G$ is the discounted return. A state abstraction function $\phi: S \rightarrow S^{'}$ aggregates states into abstract states with the goal to reduce the complexity of the state space. Given an arbitrary weighting function $w: S \rightarrow [0,1]$ s.t. $\forall s'\in S'$, $\sum_{s\in \phi^{-1}(s')}w(s)=1$, one can define an abstract reward function $R'$ and transition function $T'$ on the abstract state space $S'$:
\begin{equation}
    R'(s',a) = \sum_{s \in \phi^{-1}(s')} w(s) R(s,a) 
\end{equation}
\begin{equation}
    T'(\bar{s}'| s',a) = \sum_{\bar{s} \in \phi^{-1}(\bar{s}')} \sum_{s \in \phi^{-1}(s')} w(s) T(\bar{s}|s,a),
\end{equation}
to obtain an abstract MDP $M'=(S',A,R',T',\gamma)$. If the abstraction $\phi$ satisfies consistency criteria (see section 2, State abstraction in RL) a policy learned over $M'$ allows for optimal behaviour in $M$, which we reference from now on as base MDP \cite{li_state}. Here, we assume that we are given abstraction functions $\phi_{1},...,\phi_{n}$ with $\phi_{i}:S_{i-1} \rightarrow S_{i}$, where $S_{0}$ corresponds to the state space of the base MDP. Since $\phi_{i}$ must not necessarily satisfy any consistency criteria,  learning a policy over one of the corresponding abstract MDPs $M_{i}$ can result in a non-optimal policy. 
The goal is to learn a policy $\pi$ or action value function $Q$, that takes as input a hierarchy of state abstractions $s=(s_{1},...,s_{n})$. Here, we want to make use of the more abstract states $s_{2},...s_{n}$ for more sample efficient learning and better generalisation.

\section{Methodology}\label{sec:method}
The method can be separated into two components: (i) constructing the abstraction functions $\phi_{1},...,\phi_{n}$ from the subclass relationships in open source knowledge graphs; (ii) learning a policy over the hierarchy of abstract states $s=(s_{1},...,s_{n})$ given the abstraction functions.

\begin{figure}
    \centering
    
    \vspace{-0.25cm}
    \begin{minipage}{0.49\linewidth}
        \centering
        \includegraphics[width=\textwidth]{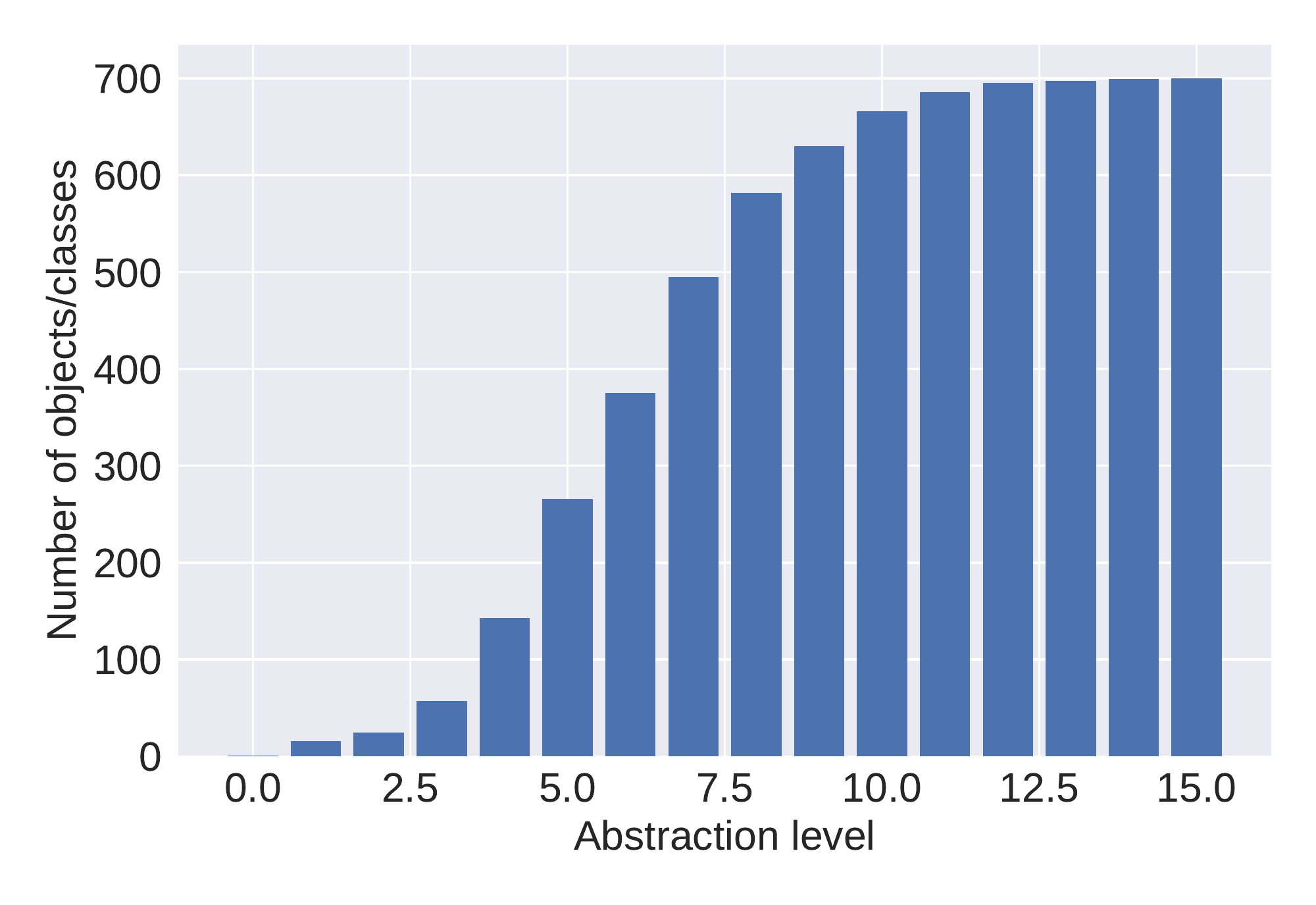} 
        \caption*{}
    \end{minipage}
    \begin{minipage}{0.49\linewidth}
        \includegraphics[width=\textwidth]{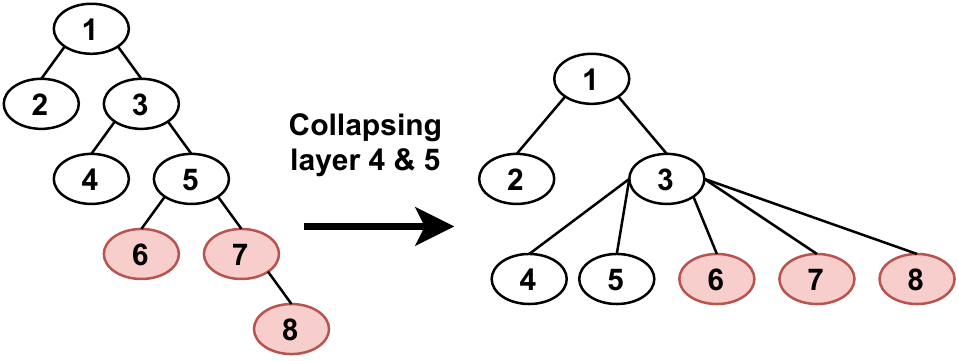} 
        \caption*{}
    \end{minipage}
    \vspace{-0.75cm}
    \caption{(Left) The number of different objects/classes that can appear for each state abstraction level in the Wordcraft environment with the superclass relation extracted from Wordnet. (Right) Visualisation of collapsing two layers in a class tree.}
    \label{fig:class_trees}
\end{figure}

\paragraph{Constructing the abstraction functions $\phi_{i}$.} A state in a commonsense game features real world entities and their relations, which can be modelled as a set, sequence or graph of entities. The idea is to replace each entity with its superclass, so that states that contain objects with the same superclass are aggregated into the same abstract state. Let $E$ be the vocabulary of symbols that can appear in any of the abstract states $s_{i}$, i.e.  $s_{i}=\{e_{1},...,e_{k} | e_{l} \in E \}$. The symbols that refer to real-world objects are denoted by $O\subseteq E$ and $C_{tree}$ represents their class tree. A class tree is a rooted tree in which the leaves are objects and the parent of each node is its superclass. The root is a generic entity class of which every object/class is a subclass (see Appendix \ref{app.A} for an example). To help define $\phi_{i}$, we introduce an entity based abstraction $\phi^{E}_{i}: E \rightarrow E$. Let $C_{k}$ represent objects/classes with depth $k$ in $C_{tree}$ and $L$ be the depth of $C_{tree}$, then we can define $\phi^{E}_{i}$ and $\phi_{i}$:
\begin{equation}
    \phi^{E}_{i}(e) = 
    \begin{cases}
    \text{Pa}(e), \mkern4mu \text{if} \mkern4mu e \in C_{L+1-i}\\
    e, \mkern4mu \text{else}, \\
    \end{cases}
\end{equation}

\begin{equation}
    \phi_{i}(s) = \{\phi^{E}_{i}(e)| e \in s\},
\end{equation}
where $\text{Pa(e)}$ are the parents of entity $e$ in the class tree $C_{tree}$
This abstraction process is visualised in Figure \ref{fig:example_abstraction}. In practice, we need to be able to extract the set of relevant objects from the game state and construct a class tree from open-source KGs. If the game state is not a set of entities but rather text, we use spaCy\footnote{https://spacy.io/} to extract all nouns as the set of objects. The class tree is extracted from either DBpedia, ConceptNet or WordNet. For the detailed algorithms of each KG extraction, we refer to Appendix \ref{app.A}. Here, we discuss some of the caveats that arise when extracting class trees from open-source KGs, and how to tackle them. Class tree can become imbalanced, i.e. the depths of the leaves, representing the objects, differs (Figure \ref{fig:class_trees}). As each additional layer with a small number of classes adds computational overhead but provides little abstraction, we collapse layers depending on their contribution towards abstraction (Figure \ref{fig:class_trees}). While in DBpedia or WordNet the found entities are mapped to a unique superclass, entities in ConceptNet are associated with multiple superclasses. To handle the case of multiple superclasses, each entity is mapped to the set of all $i$-step superclasses for $i=1,...,n$. To obtain representations for these class sets, the embeddings of each element of the set are averaged. 

\paragraph{Learning policies over a hierachy of abstract states.} 
Since prior methods in commonsense games are policy gradient-based, we will focus on this class of algorithms, while providing a similar analysis for value-based methods in Appendix \ref{app.B}. First, we look at a naive method to learn a policy in our setting, discuss its potential weaknesses and then propose a novel gradient update to overcome these weaknesses. 

A simple approach to learning a policy $\pi$ over $s$ is to take an ensemble approach by having a network with separate parameters for each abstraction level to predict logits, that are then summed up and converted via the softmax operator into a final policy $\pi$. Let $s_{i,t}$ denote the abstract state on the $i$-th level at timestep $t$, then $\pi$ is computed via:
\begin{equation}
\pi(a_{t}|s_{t}) = \textnormal{Softmax}\left(\sum_{i=1}^{n} \textnormal{NN}_{\theta_{i}}(s_{i,t})\right),
\end{equation}
where $\textnormal{NN}_{\theta_{i}}$ is a neural network processing the abstract state $s_{i}$ parameterised by $\theta_{i}$. This policy can then be trained via any policy gradient algorithm \cite{TRPO,a3c,Impala}. From here on, we will refer to this approach as sum-method.

There is no mechanism that forces the sum approach to learn on the most abstract level possible, potentially leading to worse generalisation to unseen objects. At train time making all predictions solely based on the lowest level (ignoring all higher levels) can be a solution that maximises discounted return, though it will not generalise well to unseen objects. To circumvent this problem, we adapt the policy gradient so that the parameters $\theta_{i}$ at each abstraction level are optimised to approximate an optimal policy for the $i$-th abstraction level given the computed logits on abstraction level $i-1$. Let $s^{n}_{i,t} = (s_{i,t},...,s_{n,t})$ denote the hierarchy of abstract states at timestep $t$ down to the $i$-th level. Define the policy on the $i$-th level as
\begin{equation}
    \pi_{i}(a | s^{n}_{i,t}) = \textnormal{Softmax} \left( \sum^{n}_{k=i} \textnormal{NN}_{\theta_{k}}(s_{k,t})\right).
\end{equation}
\noindent
and notice that $\pi=\pi_{1}$. To obtain a policy gradient expression that contains the abstract policies $\pi_{i}$, we write $\pi$ as a product of abstract policies:
\begin{equation}
    \pi(a | s_{t}) = \left(\prod_{i=1}^{n-1} \frac{\pi_{i}(a|s^{n}_{i,t})}{\pi_{i+1}(a|s^{n}_{i+1,t})} \right) \pi_{n}(a|s_{n,t}).
\end{equation}
and plug it into the policy gradient expression for an episodic task with discounted return $G$:
\begin{equation}\label{eq:normal_gradient}
\begin{split}
    \nabla_{\theta} J(\theta) &= \sum^{n}_{i=1} \mathbb{E}_{\pi}\left[\sum^{T}_{t=1}\nabla_{\theta} \log \left(\frac{\pi_{i}(a|s^{n}_{i,t})}{\pi_{i+1}(a|s^{n}_{i+1,t})}\right)G\right],
\end{split}
\end{equation}
where $\pi_{n+1} \equiv 1$. Notice that in Equation \ref{eq:normal_gradient}, the gradient of the parameters $\theta_{i}$ depend on the values of all policies of the level equal or lower than $i$. The idea is to take the gradient for each abstraction level $i$ only with respect to $\theta_{i}$ and not the full set of parameters $\theta$. This optimises the parameters $\theta_{i}$ not with respect to their effect on the overall policy, but their effect on the abstract policy on level $i$. The residual policy gradient is given by:
\begin{equation}\label{residual_gradient}
\begin{split}
    \nabla_{\theta} J_{res}(\theta) &= \sum^{n}_{i=1} \mathbb{E}_{\pi}\left[\sum^{T}_{t=1}\nabla_{\theta_{i}} \log(\pi_{i}(a|s^{n}_{i,t}))G\right].
\end{split}
\end{equation}
Each element of the first sum resembles the policy gradient loss of a policy over the abstract state $s_{i}$. However, the sampled trajectories are from the overall policy $\pi$ and not the policy $\pi_{i}$ and the policy $\pi_{i}$ inherits a bias from previous layers in form of logit-levels. We refer to the method based on the update in Equation \ref{residual_gradient} as residual approach. An advantage of the residual and sum approach is that the computation of the logits from each layer can be done in parallel. Any sequential processing of levels would have a prohibitively large computational overhead. 
\section{Experimental Evaluation}\label{sec:experimental_evaluation}

Our methodology is based on the assumption that abstraction via class knowledge extracted from open-source KGs is useful in commonsense game environments. This must not necessarily be true. It is first sensible to study the workings of our methodology in an idealised setting, where we control whether and how abstraction is useful for generalisation and sample efficiency. Then, we evaluate the method on two commonsense games, namely a variant of Textworld Commonsense \cite{twc} and Wordcraft.

\paragraph{Toy environment.}
We start with a rooted tree, where each node is represented via a random embedding. The leaves of the tree represent the states of a base MDP. Each depth level of the tree represents one abstraction level. Every inner node of the tree represents an abstract state that aggregates the states of its children. The optimal action for each leaf (base state) is determined by first fixing an abstraction level $l$ and randomly sampling one of five possible actions for each abstract state on that level. Then, the optimal action for a leaf is given by the sampled action for its corresponding abstract state on level $l$. The time horizon is one step, i.e. for each episode a leaf state is sampled, and if the agent chooses the correct action, it receives a reward of one. We test on new leaves with unseen random embeddings, but keep the same abstraction hierarchy and the same pattern of optimal actions. The sum and residual approach are compared to a policy trained only given the base states and a policy given only the optimal abstract states (oracle). To study the effect of noise in the abstraction, we replace the optimal action as determined by the optimal state with noise probability $\sigma$ (noise setting) or ensure that the abstraction step only aggregates a single state (ambiguity setting). All policies are trained via REINFORCE \cite{reinforce} with a value function as baseline and entropy regularisation. More details on the chosen trees and the policy network architecture can be found in Appendix \ref{app.C}.


\paragraph{Textworld Commonsense.}In text-based games, the state and action space are given as text. In Textworld Commonsense (TWC), an agent is located in a household and has to put objects in their correct commonsense location, i.e. the location one would expect these objects to be in a normal household. The agent receives a reward of one for each object that is put into the correct location. The agent is evaluated by the achieved normalised reward and the number of steps needed to solve the environment, where 50 steps is the maximum number of steps allowed. To make abstraction  necessary, we use a large number of objects per class and increase the number of games the agent sees during training from 5 to 90. This raises the exposure to different objects at training time. The agent is evaluated on a validation set where it encounters previously seen objects and a test set where it does not. The difficulty level of a game is determined by the number of rooms, the number of objects to move and the number of distractors (objects already in the correct location), which are here two, two and three respectively. To study the effect of inaccuracies in the extratcted class trees from WordNet, ConceptNet and DBpedia, we compare them to a manual aggregation of objects based on their reward and transition behaviour. Murugesan et al. \shortcite{twc} benchmark different architectures that have been proposed to solve text-based games \cite{drrn}. Here, we focus on their proposed method that makes use of a recurrent neural network architecture and numberbatch embeddings \cite{numberbatch}, which performed best in terms of normalised reward and number of steps needed. For more details on the architecture and the learning algorithm, we refer to the initial paper. As baselines, we choose: (i) adding class information via hyperbolic embeddings trained on WordNet \cite{poincare} by concatenating them to numberbatch embeddings of objects; (ii) extracting the \textit{LocatedAt} relation from ConceptNet for game objects, encoding it via graph attention and combining it with the textual embedding \cite{twc}.

\paragraph{Wordcraft.} An agent is given a goal entity and ingredient entities, and has to combine the ingredient entities to obtain the goal entity. A set of recipes determines which combination of input objects leads to which output entity. One can create different settings depending on the number of distractors (irrelevant entities in the set of recipe entities) and the set of recipes available at training time. Generalisation to unseen recipes guarantees that during training not all recipes are available, and generalisation to unseen goals ensures that at test time the goal entity has not been encountered before as a goal entity. Jiang et al. \shortcite{wordcraft} trained a policy using the IMPALA algorithm \cite{Impala}. We retain the training algorithm and the multi-head attention architecture \cite{mha} used for the policy network to train our policy components $\textnormal{NN}_{\theta_{1}},...,\textnormal{NN}_{\theta_{n}}$.

\section{Results}
After checking whether the results in the ideal environment are as expected, we discuss results in TWC and Wordcraft.
\subsection{Toy Environment}\label{sec:policy_gradient_toy}

\begin{figure}
\centering

    \begin{minipage}{\linewidth}
        \includegraphics[width=\linewidth]{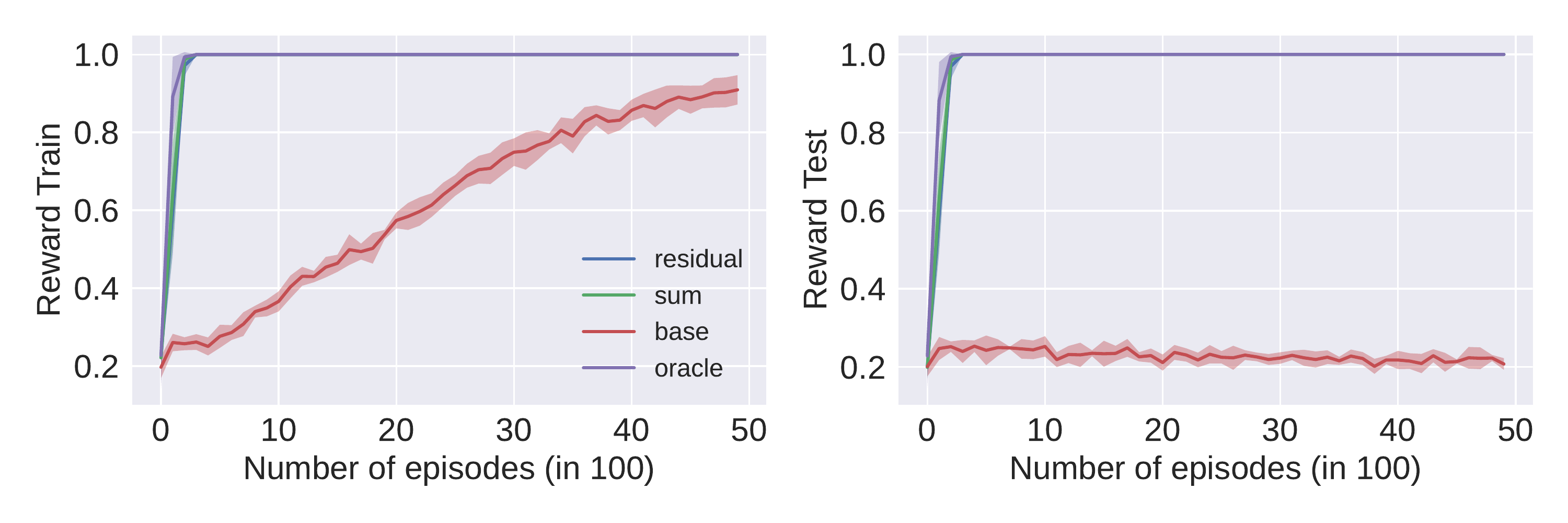} 
        \caption*{}
    \end{minipage}
    
    \vspace{-0.75cm}
    \begin{minipage}{\linewidth}
        \includegraphics[width=\linewidth]{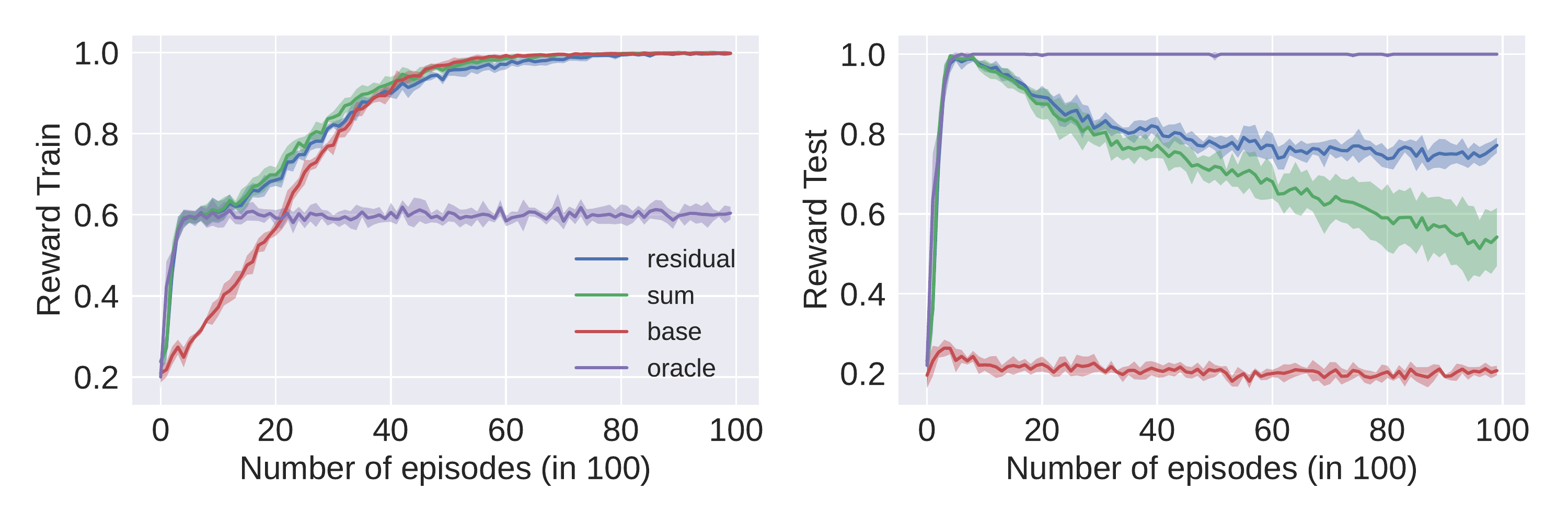} 
        \caption*{}
    \end{minipage}
    
    \vspace{-0.75cm}
    \begin{minipage}{\linewidth}
        \includegraphics[width=\linewidth]{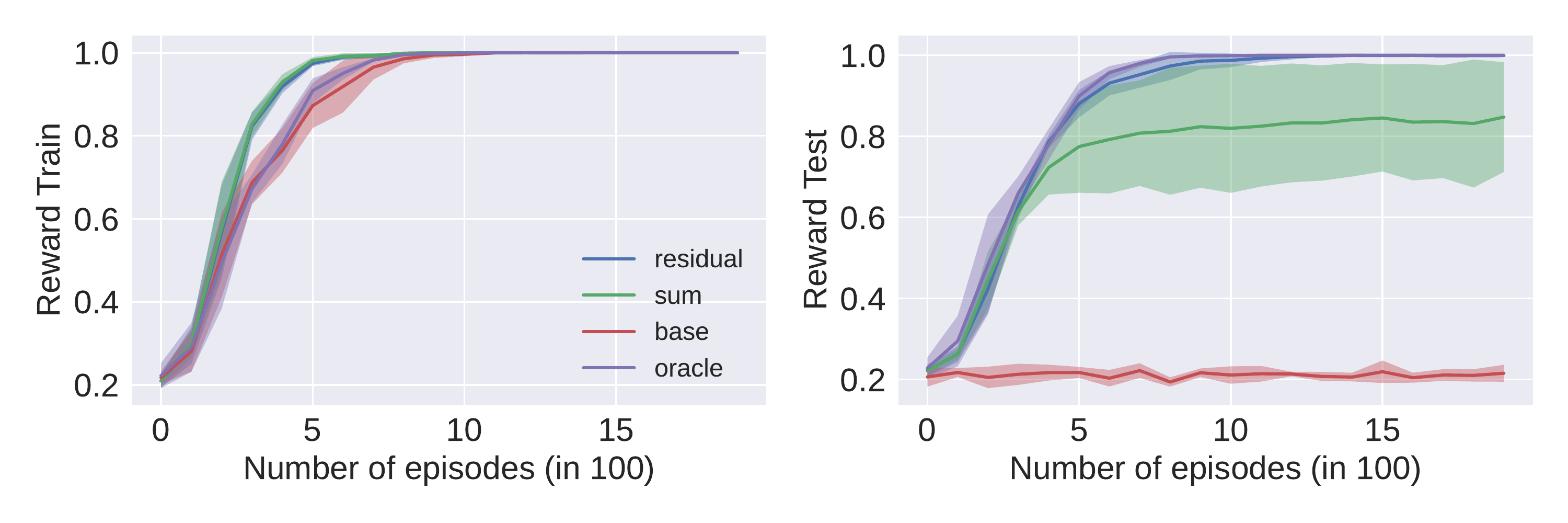} 
        \caption*{}
    \end{minipage}
    \vspace{-0.75cm}
\caption{Training and generalisation results in the toy environment. The abstract state perfectly determines the action to take in the base state (top). Noise is added to this relation, so that in 50\% of the time the optimal action is determined randomly beforehand (middle). In the ambiguous setting, not every abstract state has multiple substates (bottom). Experiments are run over five seeds.}
\label{fig:ideal_results_policy}
\end{figure}

From Figure \ref{fig:ideal_results_policy} top, one can see that the sum and residual approach are both more sample efficient than the baseline without abstraction; on par with the sample efficiency of the oracle. The same holds for the generalisation performance. As the base method faces unseen random embeddings at test time, there is no possibility for generalisation. In the noise setting (Figure \ref{fig:ideal_results_policy} middle), the policy trained on the abstract state reaches its performance limit at the percentage of states whose action is determined via their abstract state; the base, residual and sum method reach instead optimal performance. The achieved reward at generalisation time decreases for both the sum and residual method, but the sum approach suffers more from overfitting to the noise at training time when compared to the residual approach. In the case of ambiguous abstraction, we see that the residual approach outperforms the sum approach. This can be explained by the residual gradient update, which forces the residual method to learn everything on the most abstract level, while the sum approach distributes the contribution to the final policy over the abstraction layers. At test time, the sum approach puts too much emphasis on the uninformative base state causing the policy to take wrong actions. 

\subsection{Textworld Commonsense}
The results suggest that both the sum and residual approach are able to leverage the abstract states. It remains an open question whether this transfers to more complex environments with a longer time-horizon, where the abstraction is derived by extracting knowledge from an open-source KG.

\begin{figure}[h!]
\centering
\includegraphics[width=\linewidth]{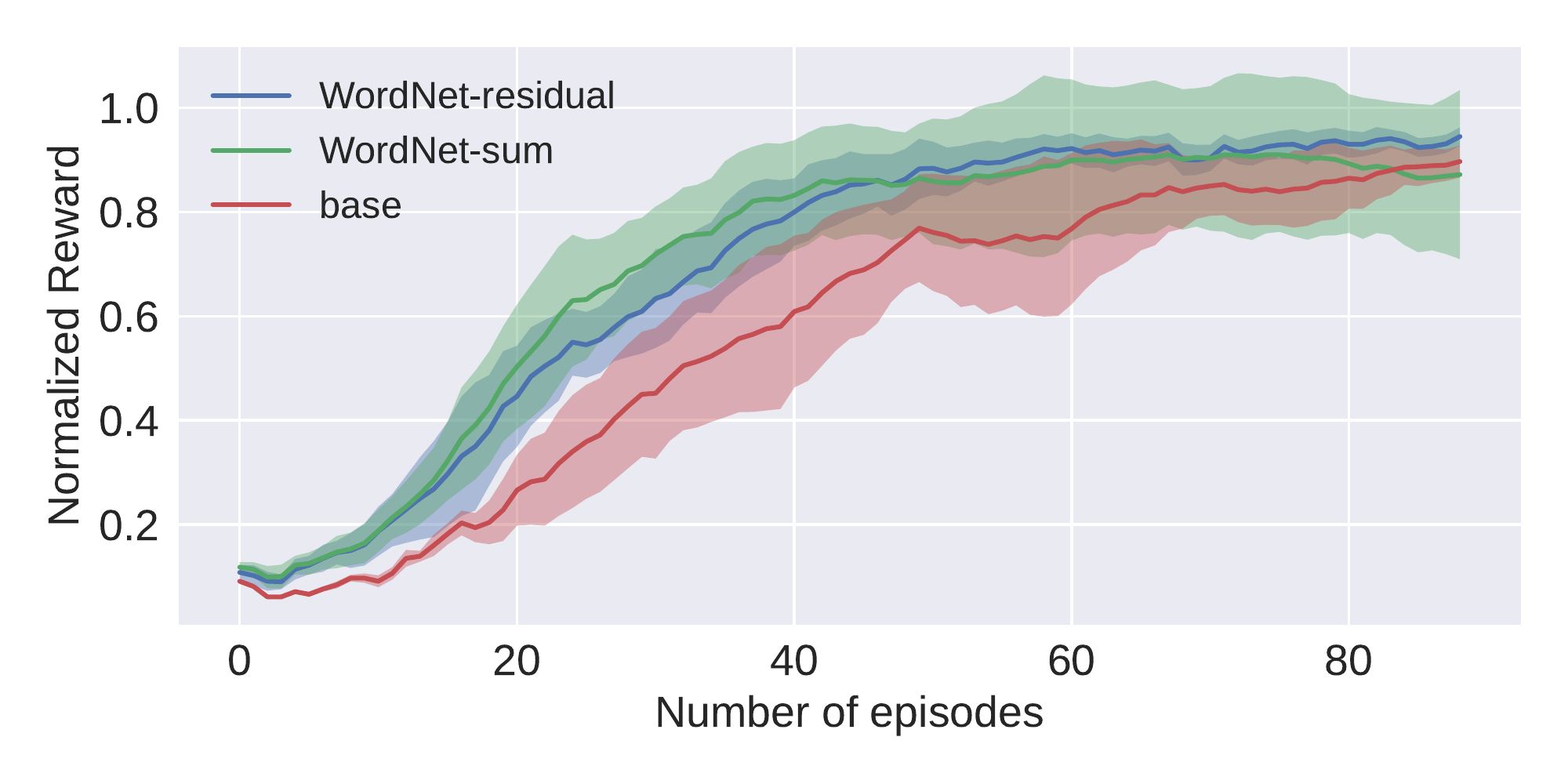} 
\caption{Training performance of the baseline (base), the sum and residual abstraction with knowledge extracted from WordNet in the difficult environment. Experiments are run over ten seeds.}
\label{fig:sample_efficiency}
\end{figure}

\paragraph{Sample efficiency.} From Figure \ref{fig:sample_efficiency}, one can see that the residual and sum approach learn with less samples compared to the base approach. The peak mean difference is 20 episodes from a total of 100 training episodes. While the sum approach learns towards the beginning of training as efficiently, it collapses for some random seeds towards the end of training, resulting in higher volatility of the achieved normalised reward. This is similar to what was previously observed in experiments in the noisy setting of the ideal environment. That the performance of the sum approach collapses for certain seeds suggests, that the learned behaviour for more abstract states depends on random initialisation of weights. This stability issue is not present for the residual approach. Figure 5 (right) shows that adding hyperbolic embeddings does not improve sample efficiency and even decreases performance towards the end of training. Adding the \textit{LocatedAt} relations improves sample efficiency to a similar degree as adding class abstractions.

\begin{table}
\centering
\setlength\tabcolsep{2.5pt}
\begin{tabular}{lllll}
\hline
\multirow{2}{*}{Method} & \multicolumn{2}{c}{Reward} & \multicolumn{2}{c}{Steps} \\
& \multicolumn{1}{c}{Valid.}  & \multicolumn{1}{c}{Test} & \multicolumn{1}{c}{Valid.} &  \multicolumn{1}{c}{Test} \\
\hline
Base &  0.91 (0.04) & 0.85 (0.05) & 25.98 (3.14) & 29.36 (3.42) \\
Base-H & 0.83 (0.06) & 0.75 (0.14) & 30.59 (5.39) & 33.92 (6.08) \\
Base-L & 0.90 (0.04) & 0.86 (0.06) & 24.83 (2.31) & 28.71 (3.46) \\
M-R & 0.96 (0.03)  & 0.96 (0.02)  & 21.26 (2.65) & 20.00 (1.57)  \\
M-S & \textbf{\textcolor{red}{0.97 (0.02)}}  & \textbf{\textcolor{red}{0.96 (0.02)}} & \textbf{\textcolor{red}{20.95 (1.38)}} & \textbf{\textcolor{red}{19.55 (1.69)}} \\
W-R &  \textbf{0.93 (0.02)} & \textbf{0.94 (0.02)} & \textbf{23.25 (1.46)}  & \textbf{24.04 (1.75)} \\
W-S &  0.88 (0.10)  & 0.87 (0.17)  & 25.69 (4.78)  & 26.21 (6.81) \\

\hline
\end{tabular}


\caption{Generalisation results for the easy and difficult level  w.r.t the validation set and the test set, in terms of mean number of steps taken (standard deviation in brackets) for each type of knowledge graph (M=manual, W=WordNet) and each approach (R=residual, S=Sum). Base refers to the baseline with no abstraction, Base-H refers to baseline with hyperbolic embeddings and Base-L for baseline added \textit{LocatedAt} relations. In bold, the best performing method without manually specified knowledge. We highlight in red the conditions in which the manual class graph outperforms the other methods. Experiments are run over ten seeds.}
\label{tab:generalisation}
\end{table}
\paragraph{Generalisation.} 
Table \ref{tab:generalisation} shows that, without any additional knowledge the performance of the baseline, measured via mean normalised reward and mean number of steps, drops when faced with unseen objects at test time. Adding hyperbolic embeddings does not alleviate that problem but rather hampers generalisation performance on the validation and test set.
When the \textit{LocatedAt} relation from ConceptNet is added to the state representation the drop in performance is reduced marginally. Given a manually defined class abstraction, generalisation to unseen objects is possible without any drop in performance for the residual and the sum approach. This remains true for the residual approach when the manually defined class knowledge is exchanged with the WordNet class knowledge. The sum approach based on WordNet performs worse on the validation and training set due to the collapse of performance at training time.


\begin{figure}
    \begin{minipage}{0.49\linewidth}
        \includegraphics[width=\linewidth]{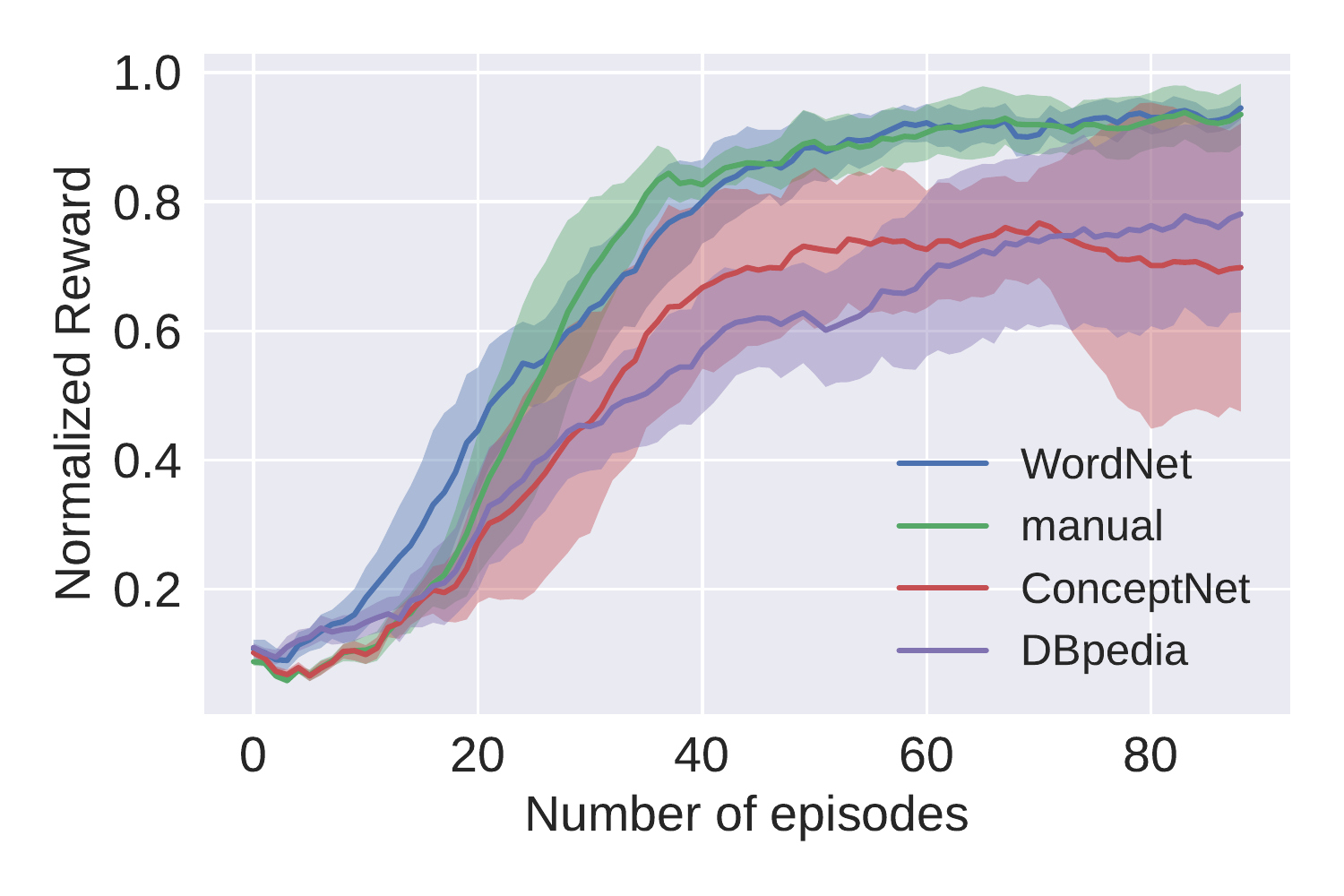} 
        \caption*{}
    \end{minipage}
    \begin{minipage}{0.49\linewidth}
        \includegraphics[width=\linewidth]{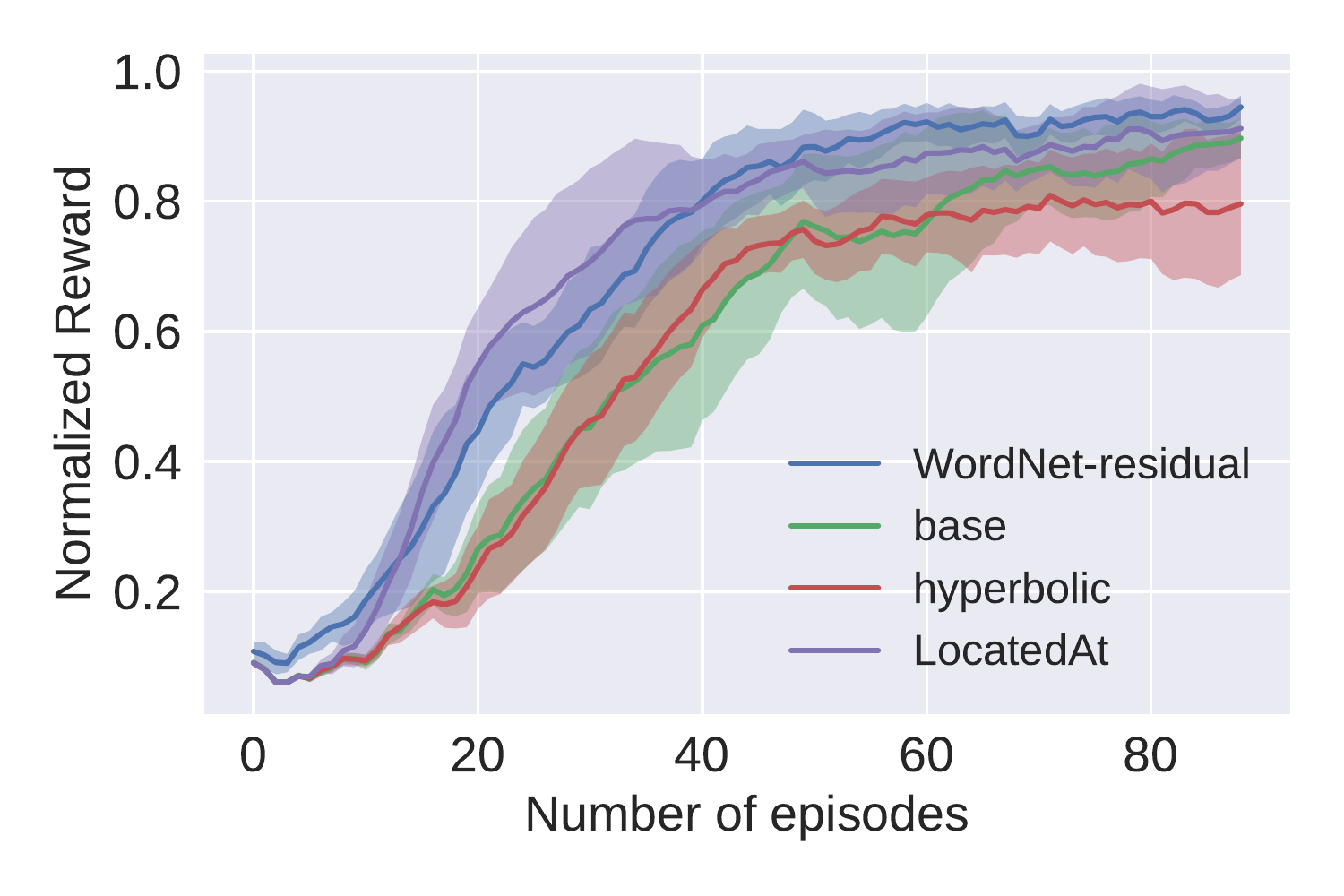} 
        \caption*{}
    \end{minipage}
    \vspace{-0.75cm}
    \caption{Training performance in the difficult environment using the residual approach when the class knowledge comes from different knowledge graphs trained (left) and when the baseline is augmented with hyperbolic embeddings or the \textit{LocatedAt} relation from ConceptNet (right). Experiments are run over ten seeds.}
    \label{fig:kg_ablation}
\end{figure}

\paragraph{KG ablation.}
From Figure \ref{fig:kg_ablation} and Table \ref{tab:generalisation} it is evident that for the residual approach, the noise and additional abstraction layers introduced by moving from the manually defined classes to the classes given by WordNet, only cause a small drop in generalisation performance and no additional sample inefficiency. The abstractions derived from DBpedia and ConceptNet hamper learning particularly for the residual approach (Figure \ref{fig:kg_ablation}). By investigating the DBpedia abstraction, we notice that many entities are not resolved correctly. Therefore objects with completely different semantics get aggregated. The poor performance from the ConceptNet abstraction hints at a problem with averaging class embeddings over multiple superclasses. Although two objects may have an overlap in their set of superclasses, the resulting embeddings could still differ heavily due to the non-overlapping classes. 

\begin{figure}
\centering
\includegraphics[width=\linewidth]{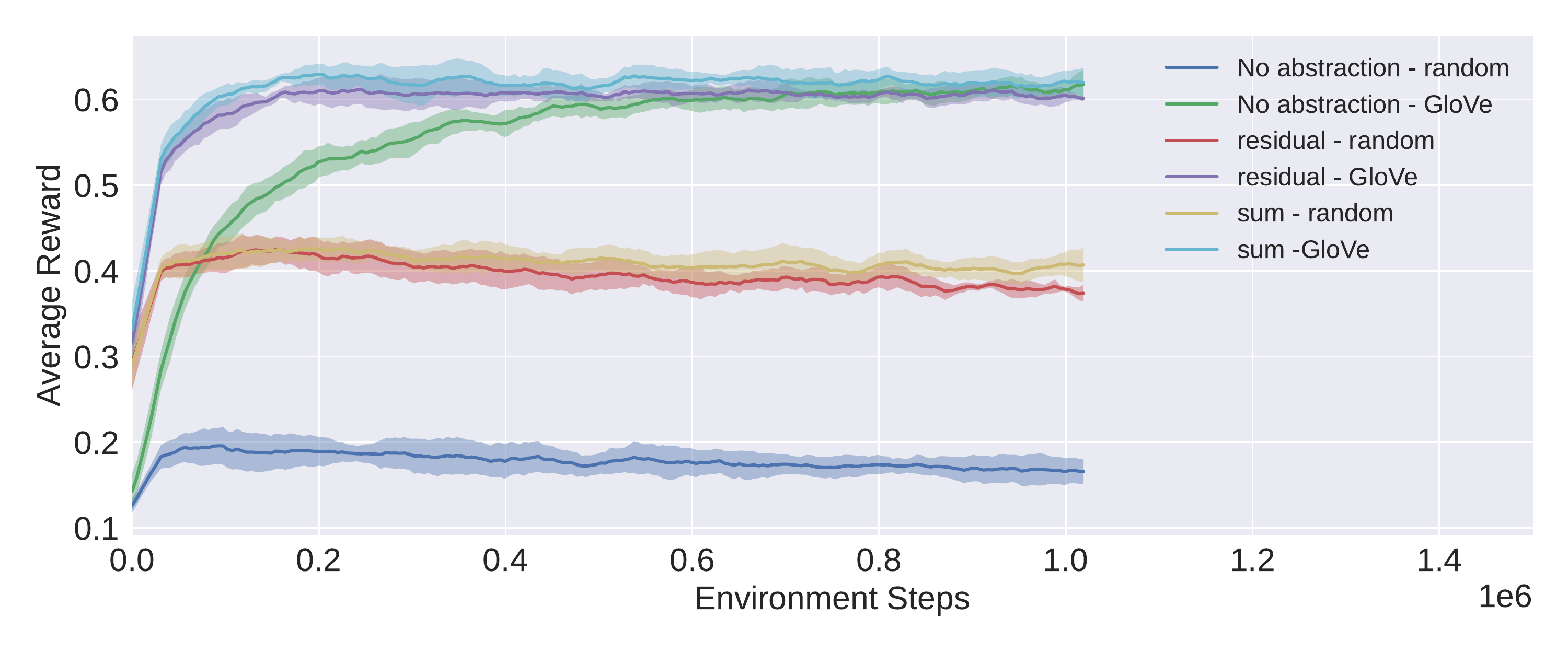} 
\caption{Generalisation results for the Wordcraft environment with respect to unseen goal entities. Experiments are run over ten seeds.}
\label{fig:wordcraft}
\end{figure}

\subsection{Wordcraft}\label{sec:wordcraft_results}
Figure \ref{fig:wordcraft} shows the generalisation performance in the Wordcraft environment. For the case of random embeddings, abstraction can help to improve generalisation results. Replacing random embeddings with GloVe embeddings improves generalisation beyond the class abstraction, and combining class abstraction with GloVe embeddings does not result in any additional benefit. Looking at the recipe structure in Wordcraft, objects are combined based on their semantic similarity, which can be better captured via word embeddings rather than through classes. Although no generalisation gains can be identified, adding abstraction in the presence of GloVe embeddings leads to improved sample efficiency. A more detailed analysis discussing the difference in generalisation between class abstraction and pretrained embeddings in the Wordcraft environment can be found in Appendix \ref{app.E}.

\section{Conclusion}
We show how class knowledge, extracted from open-source KGs, can be leveraged to learn behaviour for classes instead of individual objects in commonsense games. To force the RL agent to make use of class knowledge even in the presence of noise, we propose a novel residual policy gradient update based on an ensemble learning approach. If the extracted class structure approximates relevant classes in the environment, the sample efficiency and generalisation performance to unseen objects are improved. Future work could look at other settings where imperfect prior knowledge is given and whether this can be leveraged for learning. Finally, KGs contain more semantic information than classes, which future work could try to leverage for an RL agent in commonsense games.

\section*{Acknowledgements}
This research was (partially) funded by the Hybrid Intelligence Center, a 10-year programme funded by the Dutch Ministry of Education, Culture and Science through the Netherlands Organisation for Scientific Research, https://hybrid-intelligence-centre.nl.


{\bibliographystyle{named}
\small
\bibliography{ijcai22}}

\newpage
\appendix

\section{Extracting class trees from commonsense KGs} \label{app.A}
To construct the abstraction function $\phi_{i}$ presented in Section \ref{sec:method} from open-source knowledge graphs, we need a list of objects that can appear in the game. We collect trajectories via random play and extract the objects out of each state appearing in these trajectories. The extraction method depends on the representation of the state. In Wordcraft \cite{wordcraft} (see Section \ref{sec:experimental_evaluation}), the state is a set of objects, such that no extraction is needed. In Textworld \cite{textworld} the state consists of text. We use spaCy\footnote{https://spacy.io/} for POS-tagging and take all nouns to be relevant objects in the game. Next, we look at how to extract class knowledge from WordNet \cite{wordnet}, DBpedia \cite{dbpedia} and ConceptNet \cite{conceptnet} given a set of relevant objects. An overview of the characteristics of the extracted class trees for each knowledge graph and each environment is presented in Table \ref{tab:class_tree_characteristics}.

\begin{table}[h!]
    \centering
    \begin{tabular}{llll}
        Env. & KG & Miss. Ent. & \#Layers   \\
        \hline
        \multirow{3}{*}{TWC} & WordNet & 2/148 & 9 \\
        & DBpedia & 2/148 & 6\\
        & ConceptNet & 6/148 & 2 \\
        
        \hline
        Wordcraft  & WordNet & 13/700 & 16 \\

    \end{tabular}
    \caption{Characteristics of the extracted class trees for each KG in the difficult TWC games and Wordcraft. Missing entities (Miss. Ent.) represents the number of entities that could not be matched to an entity in the KG, relative to the total number of game entities. The number of layers (\#Layers) does not include the base level.}
    \label{tab:class_tree_characteristics}
\end{table}

\subsection{WordNet}\label{WN}

\begin{figure*}
\centering
\includegraphics[width=\linewidth]{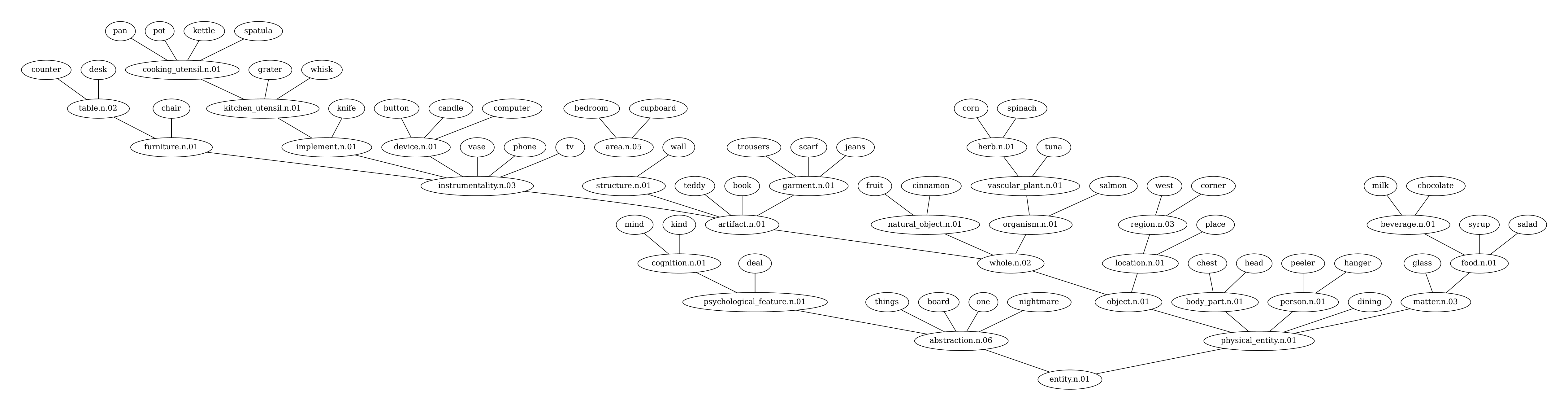} 
\caption{Visualisation of the class tree constructed from a subset of entities appearing in the Textworld Commonsense environment. The subclass relations are extracted from WordNet.}
\label{fig:class_tree}
\end{figure*}

We use WordNet's hypernym relation between two nouns as the superclass relation. To query WordNet's synsets and the hypernym relation between them, we use NLTK\footnote{https://www.nltk.org/}. For each object, we query recursively the superclass relation until the entity synset is reached. Having obtained this chain of superclasses for each object, we merge it into a class tree by aggregating superclasses appearing in multiple superclass chains of different objects into one node. An example of the resulting class tree can be found in Figure \ref{fig:class_tree}.
One problem that arises is entity resolution, i.e. picking the semantically correct synset for a given word in case of multiple available semantics. Here, we simply take the first synset/hypernym returned. More complex resolution methods are possible. Objects that have no corresponding synsets in WordNet have the noun "entity" as a parent in the class tree. 

\subsection{DBpedia} 

We use the SPARQL \footnote{https://www.w3.org/TR/rdf-sparql-query/} language to first query all the URI's that have the objects name as a label (using the rdfs:label property). If a URI that is part of DBpedia's ontology can be found, we query the \textit{rdfs:subClassOf} relation recursively until the URI \textit{owl:Thing} is reached. If the URI corresponding to the object name is not part of the ontology, we query the \textit{rdf:type} relation recursively until the URI \textit{owl:Thing} is reached. This gives us again a chain of classes connected via the superclass relation for each object, which can be merged into a class tree as described in Section A.1. Any entity that can not be mapped to a URI in DBpedia is directly mapped to the \textit{owl:Thing} node of the class tree. 

\subsection{ConceptNet}

We use ConceptNet's API\footnote{https://conceptnet.io/} to query the \textit{IsA} relation for a list of objects. In ConceptNet, it can happen that there are multiple nodes with different descriptions for the same semantic concept connected to an object via the \textit{IsA} relation. For example, the object airplane is connected among other things via the \textit{IsA} relation to \textit{heavier-than-air craft}, \textit{aircraft} and \textit{fixed-wing aircraft}. This causes two problems: (i) sampling only one superclass could lead to a class tree where objects of the same superclass are not aggregated due to minor differences in the class description; (ii) recursively querying the superclass relation until a general entity class is reached leads to a prohibitively large class tree, and is not guaranteed to terminate due to possible cycles in the \textit{IsA} relation. To tackle the first problem, we maintain a set of superclasses for each object per abstraction level, instead of a single superclass per level. The representative embedding of the superclass of an object is then calculated by averaging the embeddings of the elements in the set of superclasses. To tackle the second problem, we restrict the superclasses to the two step neighbourhood of the \textit{IsA} relation, i.e. we have two abstraction levels. Further simplifications are that only \textit{IsA} relations are considered with a weight above or equal to one\footnote{The weight of a relation in ConceptNet is a measure of confidence of whether the relation is correct.} and superclasses with a description length of less than or equal to three words. 

\section{Residual Q-learning} \label{app.B}
First, the adaption of Q-learning to the setting described in section \ref{sec:problem_formulation} is explained. Then results of the algorithm in the toy environment from section \ref{sec:experimental_evaluation} are presented.

\subsection{Algorithm}

In Section \ref{sec:method} we focused on how to adapt policy gradient methods to learn a policy over a hierarchy of abstract states $s=(s_{1},...,s_{n})$. Here we provide a similar adaption of Q-learning \cite{q_learning} and present results from the toy environment. The sum approach naturally extends to Q-learning but instead of summing over logits from each layer, we sum over action values. Let $s_{t}=(s_{1,t},...,s_{n,t})$ be the hierarchy of abstract states at time $t$, then the action value is given by:
\begin{equation}
    Q(s_{t},a) = \sum_{i=1}^{n} NN_{\theta_{i}}(s_{i},a),
\end{equation}
where $NN_{\theta_{i}}$ is a neural network parameterized by $\theta_{i}$. The idea of the residual Q-learning method is similar to the residual policy-gradient, i.e. we want to learn an action value function on the most abstract level and then adapt this on lower levels as more information on the state becomes available. Let $s_{t}=(s_{1,t},...,s_{n,t})$ be a hierarchy of abstract states at time $t$ and $R_{i}$ the reward function of the abstract MDP on the i-th level as defined in Section \ref{sec:problem_formulation}. Further define $R_{n+1} \equiv 0$. Then one can write the action value function for an episodic task with discount factor $\gamma$ as:
\begin{equation*}
\begin{split}
    Q(s,a) &= \mathbb{E}[\sum_{t=0}^{T} \gamma^{t} R_{1,t}|S=s,A=a] \\
     & =\mathbb{E}[\sum_{t=0}^{T} \gamma^{t}\sum_{i=1}^{n} \left(R_{i,t}- R_{i+1,t}\right)|S=s,A=a] \\
     & = \sum_{i=1}^{n} Q_{i}^{res}(s_{i,t},s_{i+1,t},a), \\ 
\end{split}
\end{equation*}
where $Q_{i}^{res}$ is defined as:
\begin{equation*}
Q_{i}^{res}(s_{i,t},s_{i+1,t},a)=\mathbb{E}[\sum\limits_{t=0}^{T} \gamma^{t}\left(R_{i,t}- R_{i+1,t}\right)|S=s,A=a].
\end{equation*}
The dependency of the reward function $R_{i,t}$ on the state $s_{i,t}$ and action $a$ is omitted for brevity.
At each layer of the abstraction the residual action value function $Q_{i}^{res}$ learns how to adapt the action values of the previous layer to approximate the discounted return for the now more detailed state-action pair. This information difference is characterised by the difference in the reward functions of both layers $R_{i}-R_{i+1}$.

To learn $Q_{i}^{res}$ we use an adapted version of the Q-learning update rule. Given a transition $(s_{t},a,r_{t},s_{t+1})$, where $s$ represents the full hierarchy over abstract states, the update is given by:
\begin{equation*}
Q_{i}^{res} = (1-\alpha) Q^{res}_{i} + \alpha(R_{i}-R_{i+1} + \gamma \max_{a \in A}Q_{i}^{res})
\end{equation*}
This can be seen as performing Q-learning over the following residual MDP $M^{res}_{i}=(S_{i}\times S_{i+1},A,R_{i}^{res},T_{i}^{res},\gamma)$ with
\begin{equation*}
\begin{split}
    &R^{res}_{i} = R_{i}(s_{i,t},a) - R_{i+1}(s_{i+1,t},a) \quad \text{and}\\
    &T^{res}_{i} = T_{i}(s_{i,t+1} | s_{i,t},a) \cdot T_{i+1}(s_{i+1,t+1} | s_{i+1,t},a).
\end{split}
\end{equation*}
We cannot apply the update rule directly, since each transition only contains a single sample for both reward functions, i.e. $r_{t}$ is a sample of $R_{i}$ and $R_{i+1}$. As a result for every transition the difference would be 0. To be able to apply the learning rule, one solution would be to maintain averages of the reward functions for each state. This is infeasible in high-dimensional state spaces. Instead, we approximate $R_{i+1}$ via the difference between the current and next state action value of the previous abstraction level, i.e.
\begin{equation}
    R_{i+1} = Q_{i+1}(s_{i+1,t},a)-\gamma Q_{i+1}(s_{i+1,t+1},a).
\end{equation}
The adaptions made by Mnih et al. \shortcite{mnih_atari} to leverage deep neural networks for Q-learning can be used to scale residual Q-learning to high-dimensional state spaces. 

\begin{figure}
\centering

    \begin{minipage}{\linewidth}
        \includegraphics[width=\linewidth]{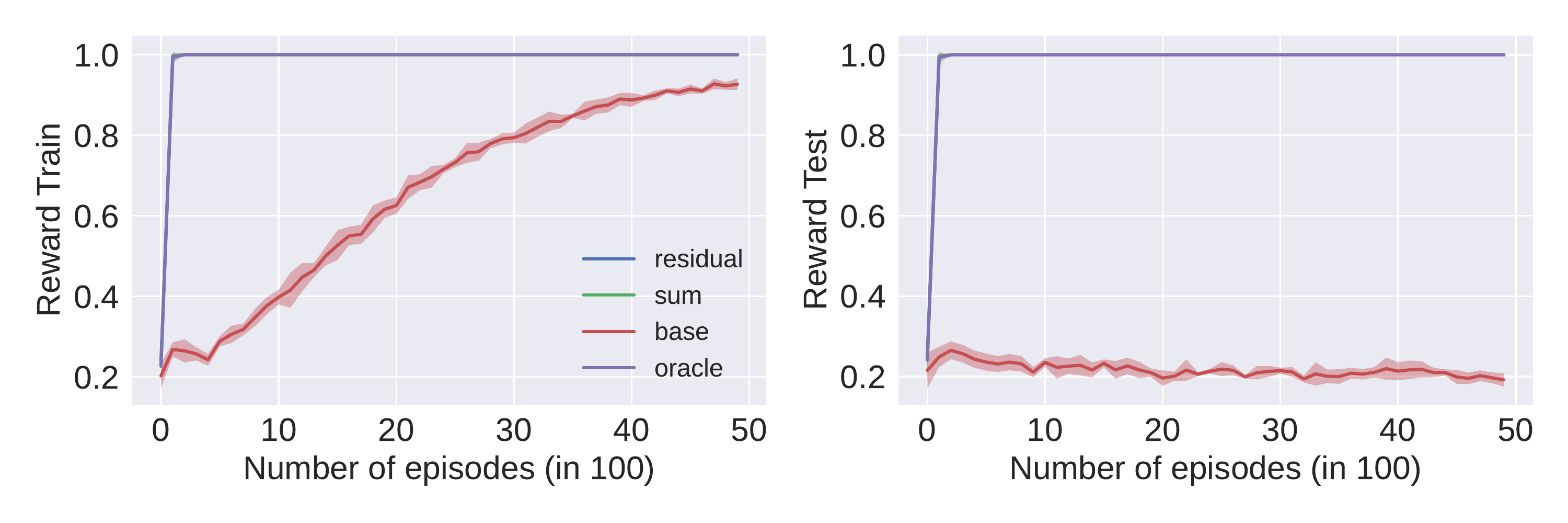} 
        \caption*{}
    \end{minipage}
    
    \vspace{-0.75cm}
    \begin{minipage}{\linewidth}
        \includegraphics[width=\linewidth]{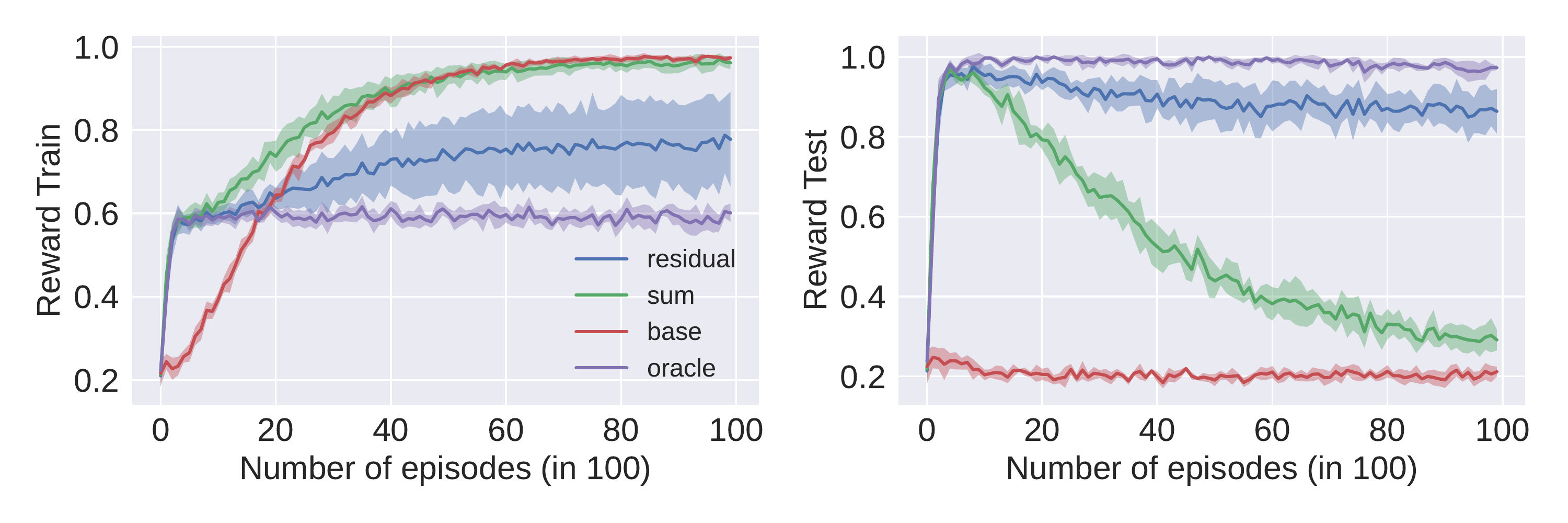} 
        \caption*{}
    \end{minipage}
    
    \vspace{-0.75cm}
    \begin{minipage}{\linewidth}
        \includegraphics[width=\linewidth]{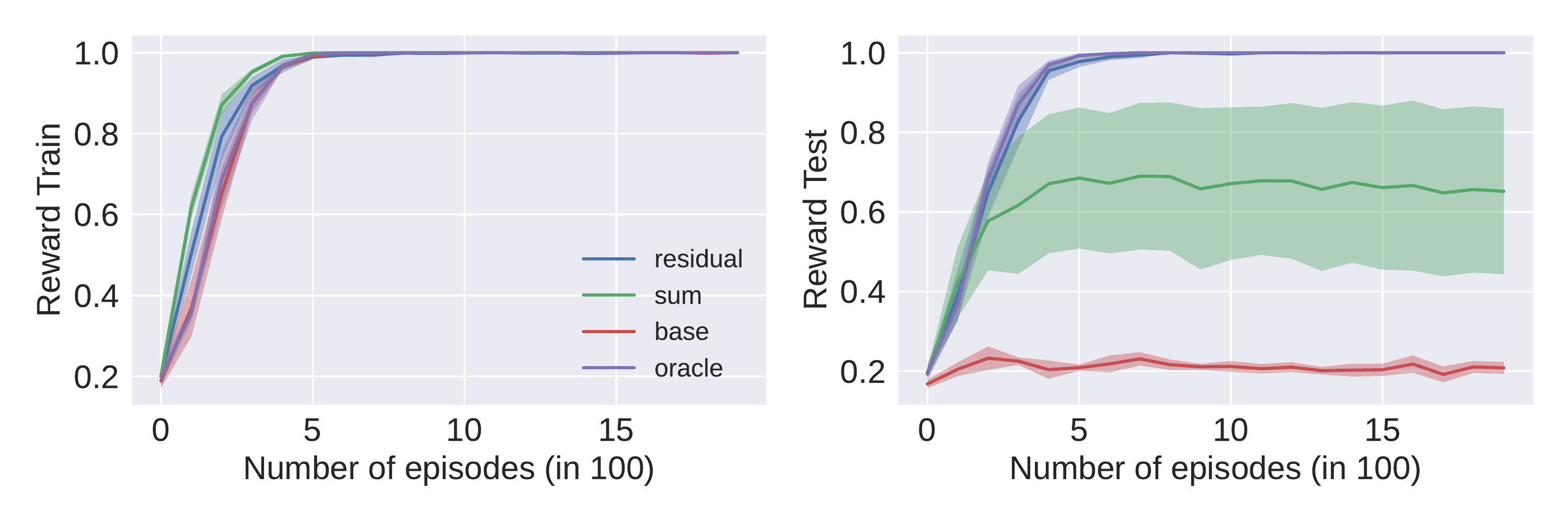} 
        \caption*{}
    \end{minipage}
    \vspace{-0.75cm}
\caption{Training and generalisation results in the ideal environment. The abstract state perfectly determines the action to take in the base state (top). Noise is added to this relation, such that in 40\% of the time the optimal action is determined randomly beforehand (middle). In the ambiguous setting not every abstract state has multiple substates (bottom). Experiments are run over five seeds.}
\label{fig:ideal_results_dqn}
\end{figure}

\subsection{Results in the Toy Environment}

The results resemble the results already discussed for the policy gradient case (cfr. Section \ref{sec:policy_gradient_toy}). Without any noise or ambiguity in the state abstraction the residual and sum approach learn as efficiently with respect to the number of samples and generalise as well as if the perfect abstraction is given (Figure \ref{fig:ideal_results_dqn} top). The sum approach struggles to learn the action values on the correct abstraction level if the abstraction is ambiguous, hampering its generalisation performance (Figure \ref{fig:ideal_results_dqn} bottom). Both, the sum and residual approach, generalise worse in the presence of noise (Figure \ref{fig:ideal_results_dqn} middle). However, the achieved reward at test time is less influenced by noise for the residual approach. We did not evaluate the residual Q-learning approach on the more complex environments as policy gradient methods have shown better performance in previous work \cite{twc} 

\section{Toy environment} \label{app.C}
Here, a visualisation of the toy environment, details on the tree representing the environment and the used network architecture for the policy are given.

\subsection{Environment Details}
As explained in Section \ref{sec:experimental_evaluation} the environment is determined by a tree where each node represents a state of the base MDP (leaves) or an abstract state (inner nodes). In a noise free setting the optimal action for a base state is determined by a corresponding abstract state, while with noise the optimal action is sampled at random with probability $\sigma$ or else determined by the corresponding abstract state. 

\begin{figure}[h!]
\begin{minipage}{\linewidth}
    \centering
    \includegraphics[width=\linewidth]{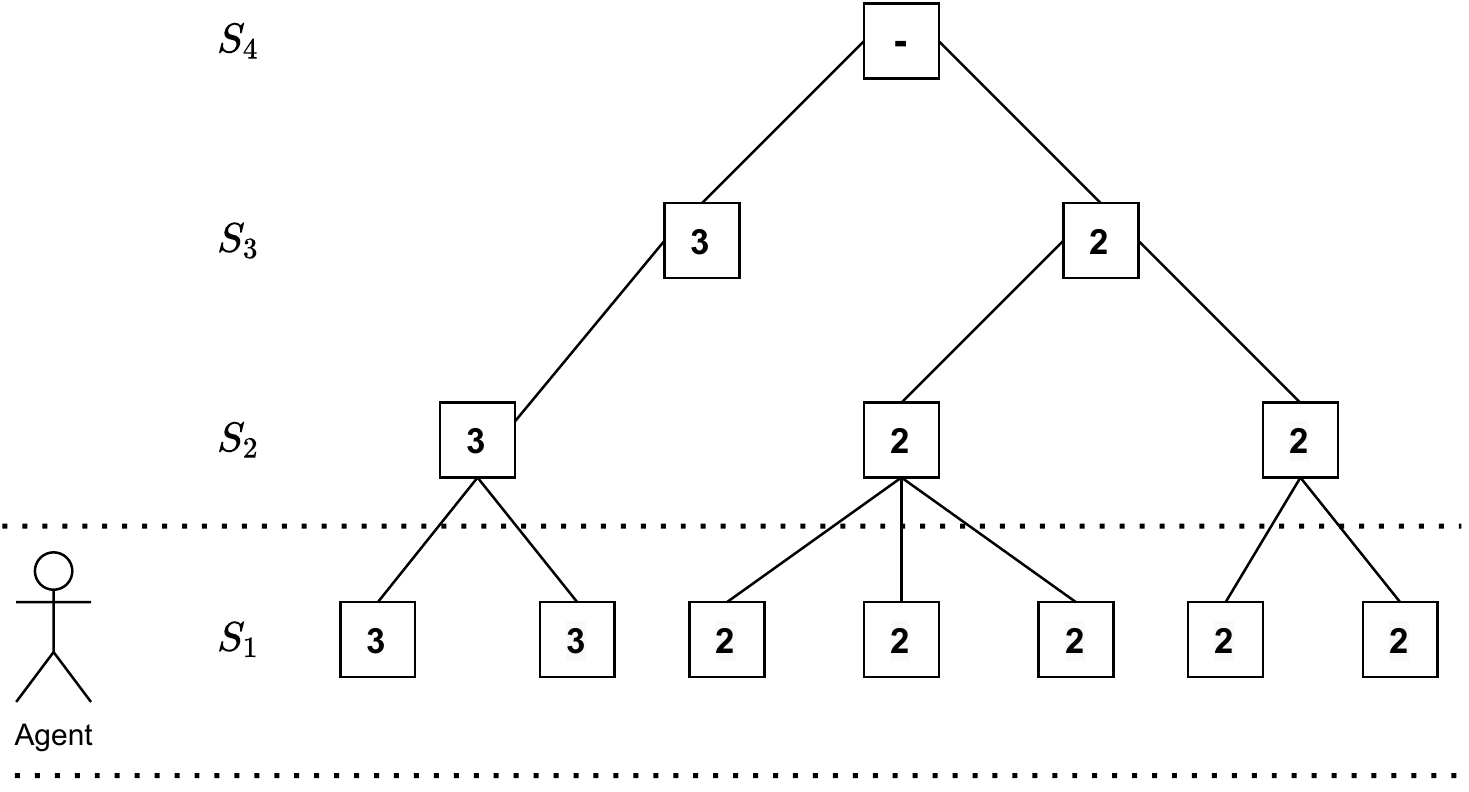}
    \caption*{}
\end{minipage}
\begin{minipage}{\linewidth}
    \centering
    \includegraphics[width=\linewidth]{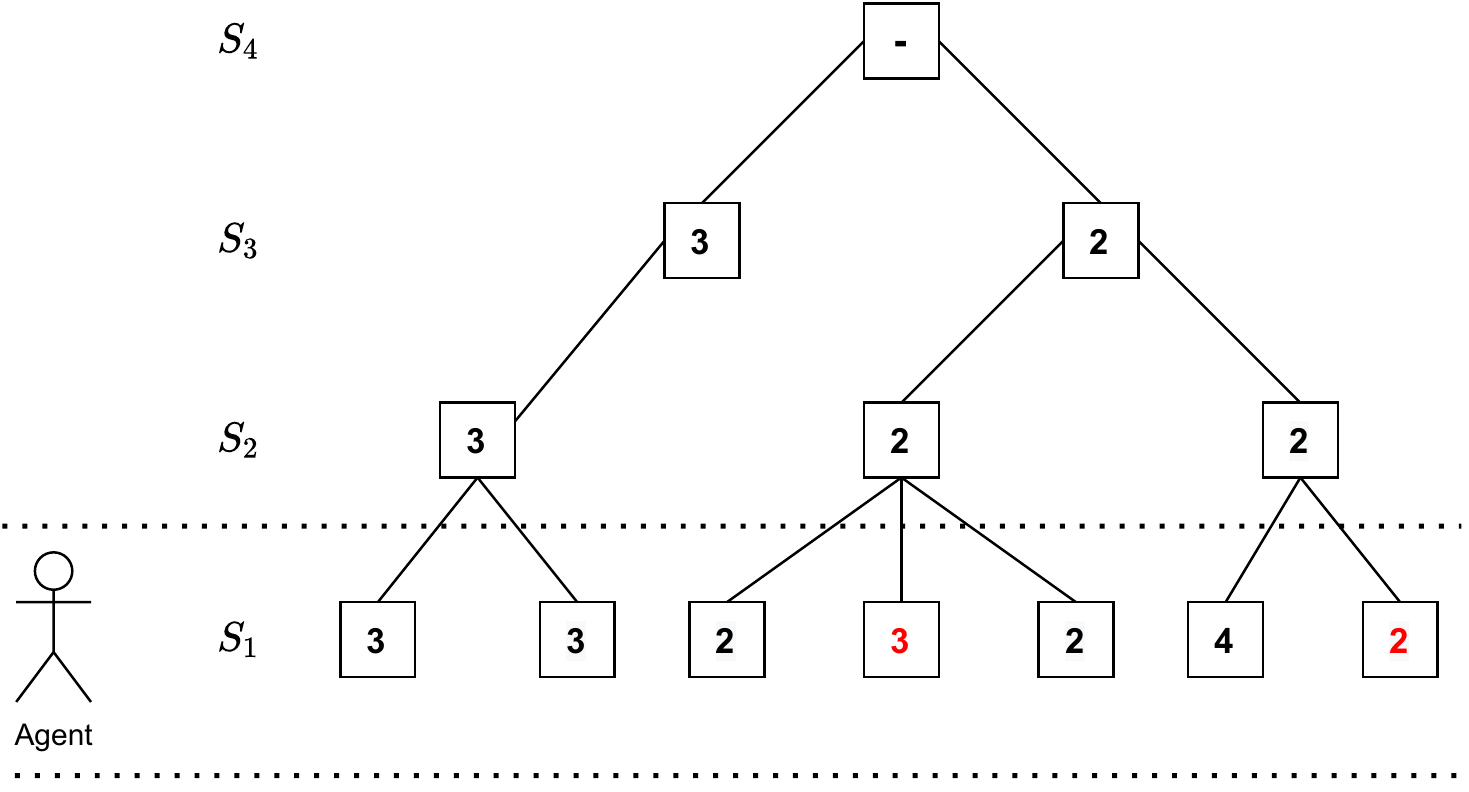}
    \caption*{}
\end{minipage}
\begin{minipage}{\linewidth}
    \centering
    \includegraphics[width=\linewidth]{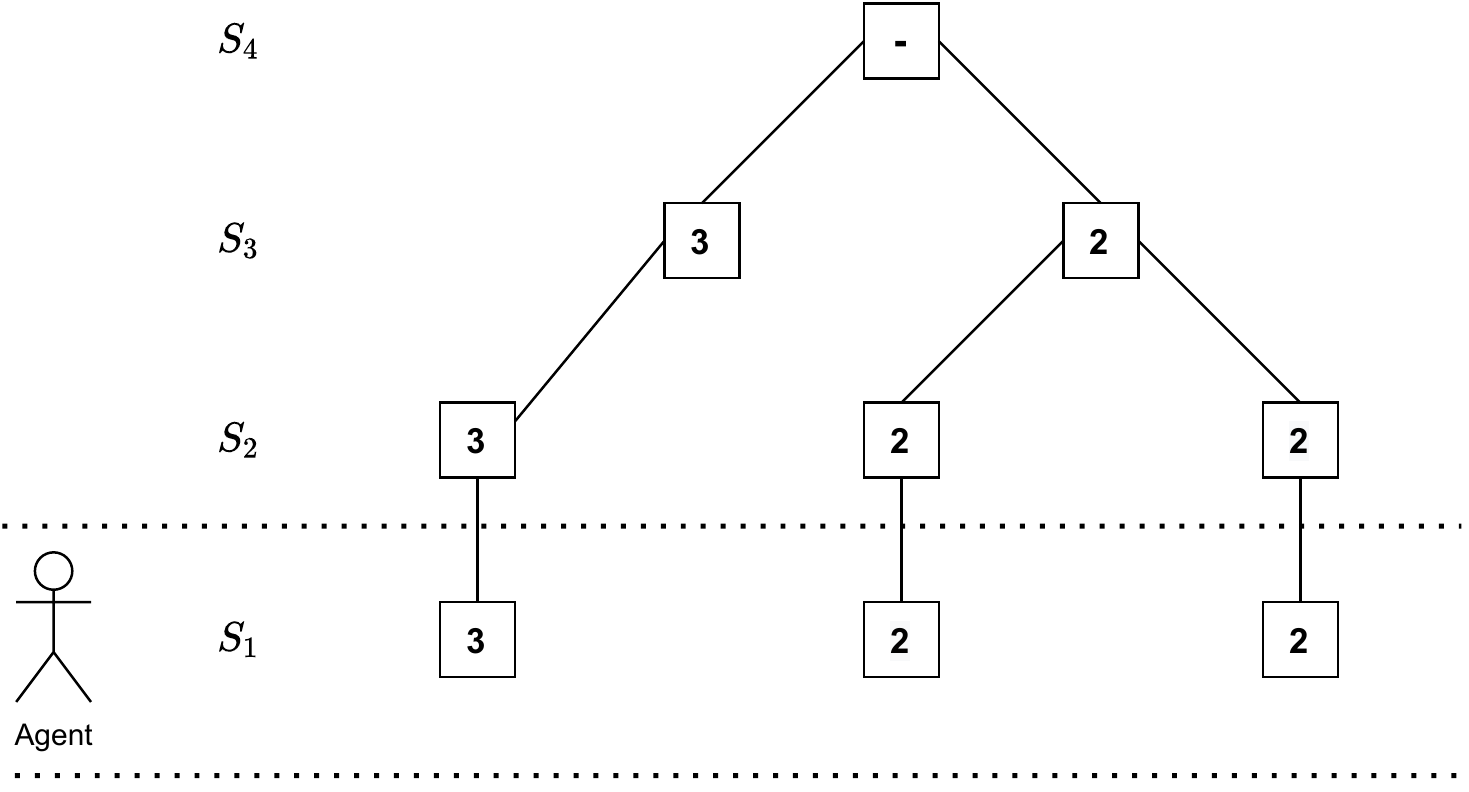}
\end{minipage}
\caption{Simplified version of the tree used to define the toy environment. In the base method the agent can only observe the leaf states. The optimal action for each state is displayed by the number in each node. At top the environment in the basic setting is displayed, in the middle with noise and at the bottom with ambiguous abstraction.}
\label{fig:example_toy_environment}
\end{figure}

A visualisation of the environment and what changes for each condition can be seen in Figure \ref{fig:example_toy_environment} and is summarised in Table \ref{tab:toy_environment}.

\begin{table}[h!]
    \centering
    \begin{tabular}{llll}
        Setting & Branching factor & $\sigma$ & Abstraction ambiguity \\
        \hline
        Basic & [7,10,8,8] & 0 & No \\
        Noise & [7,10,8,8] & 0.5 & No \\
        Ambig. & [7,10,8,1] & 0 & Yes
    \end{tabular}
    \caption{Properties of the environment tree for the three settings: basic, noise and ambiguous with noise probability $\sigma$.}
    \label{tab:toy_environment}
\end{table}

\subsection{Network architecture and Hyperparameter}

The policy is trained via REINFORCE \cite{reinforce} with a value function as baseline and entropy regularisation. The value function and policy are parameterised by two separate networks and the entropy regularisation parameter is $\beta$. For the baseline and the optimal abstraction method the value network is a one layer MLP of size 128 and the policy is a three-layer MLP of size $(256,128,64)$. In case abstraction is available the value network is kept the same but the policy network is duplicated (with a separate set of weights) for each abstraction level to parameterise the $NN_{\theta_{i}}$ (Section \ref{sec:method}). 

\begin{table}[h!]
    \centering
\begin{tabular}{ll}
    Hyperparameter & Value  \\
    \hline 
    number of episodes & 5000 \\
    evaluation frequency & 100 \\ 
    number of evaluation samples & 20 \\
    batch size & 32 \\
    state embedding size & 300  \\
    learning rate & 0.0001 \\
    $\beta$ & 1 
\end{tabular}
    \caption{Hyperparameters for training in the toy environment.}
    \label{tab:my_label}
\end{table}

\section{Wordcraft environment} \label{app.E}
\subsection{Network architecture and Hyperparameter}
We build upon the network architecture and training algorithm used by Jiang et al. \shortcite{wordcraft}. Here we only present the changes necessary to compute the abstract policies $\pi_{i}$ defined in Section \ref{sec:method} and the chosen hyperparameters.

To process a hierarchy of states $s_{t} = (s_{t,1},...,s_{t,n})$ one can simply add a dimension on the state tensor representing the abstraction level. Classical network layers such as multi-layer perceptron (MLP) can be adapted to process such a state tensor by adding a weight dimension representing the weights for each abstraction level and using batched matrix multiplication. We tuned the learning rate and entropy regularisation parameter ($\beta$) by performing a grid search over possible parameter values. We decided to collapse layers based on the additional abstraction each layer contributed (Section \ref{sec:method}). For the sum and residual approach a learning rate of $0.0001$ and $\beta=0.1$ were used across experiments and layer 9-15 of the class hierarchy were collapsed. For the baseline a learning rate of $0.001$ and the $\beta=0.1$ were used. Only in the unseen goal condition the baseline with Glove embedding had to be trained with a learning rate of $0.0001$.

\subsection{Additional Results}

\begin{figure}
\centering
\includegraphics[width=\linewidth]{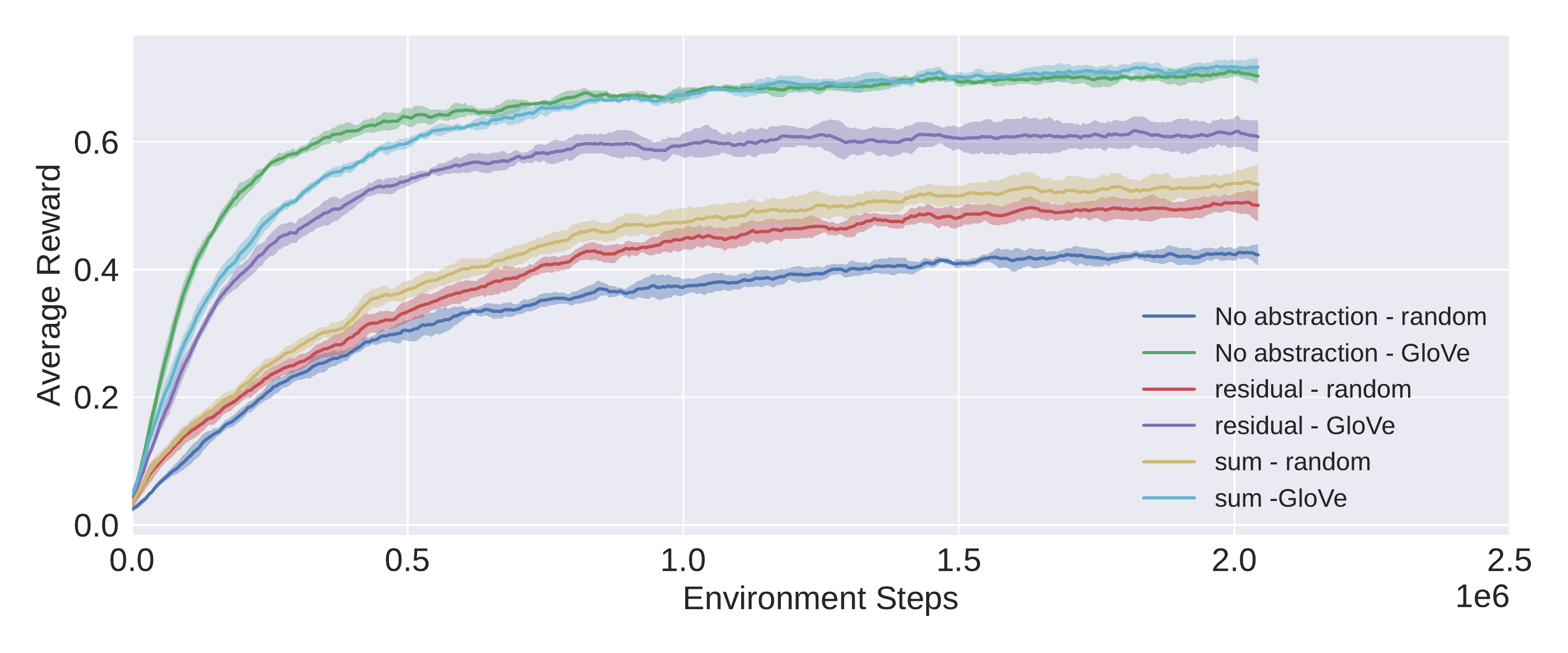} 
\caption{Generalisation results for the Wordcraft environment with respect to unseen recipes. Experiments are run over 5 seeds.}
\label{fig:wordcraft_unseen_recipes}
\end{figure}

In the Wordcraft environment the results with respect to generalisation to unseen recipes (Figure \ref{fig:wordcraft_unseen_recipes}) show similar patterns to the generalisation results presented in Section \ref{sec:wordcraft_results}. With random embeddings adding class abstraction improves generalisation, but not as much if random embeddings are replaced with Glove embeddings \cite{glove}. The combination of class abstraction with Glove embeddings does not lead to additional improvements. The objects that need to be combined in Wordcraft rarely follow any class rules, i.e. objects are not combined based on their class. Word embeddings capture more fine-grained semantics rather than just class knowledge \cite{glove}, therefore it is possible that class prior knowledge is less prone to overfitting. In a low data setting it is possible that word embeddings contain spurious correlations which a policy can leverage to perform well on during training but do not generalize well.

\begin{figure}
\centering
\includegraphics[width=\linewidth]{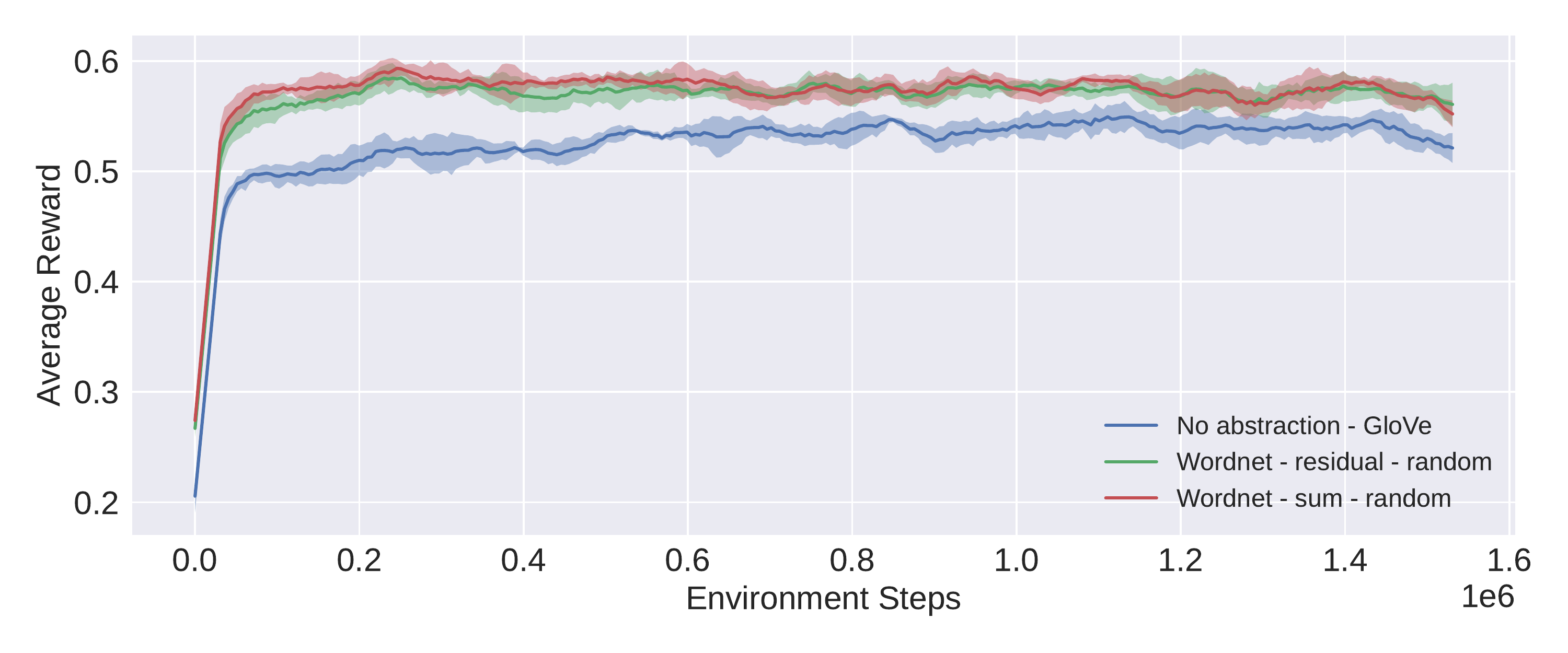} 
\caption{Generalisation results for the Wordcraft environment with respect to unseen goal entities trained based on 10\% of all available recipes instead of 80\%. Experiments are run over 5 seeds.}
\label{fig:wordcraft_low_data}
\end{figure}

To test this hypothesis we created a low data setting in which training was performed when only 10\% of all possible goal entities were part of the training set instead of the usual 80\%. In Figure \ref{fig:wordcraft_low_data} one can see that all methods have a drop in performance (cfr. Section \ref{sec:wordcraft_results}), but the drop is substanially smaller for the methods with abstraction and random embeddings compared to the baseline with Glove embeddings. The generalisation performance reverses, i.e. the baseline with Glove embeddings now performs the worst.

To understand whether and how the learned policy makes use of the class hierarchy we visualise a policy trained with the residual approach, random embeddings and access to the the full class hierarchy extracted from WordNet.
For the visualisation we choose recipes that follow a class rule. In Wordcraft one can create an airplane by combining a bird and something that is close to a machine. From Figure \ref{fig:policy_visualisation} one can see that the policy learns to pick the correct action on level nine by recognising that a vertebrate and machine need to be combined.


\section{Runtime Analysis}\label{app.F}

\begin{figure}
\centering
\includegraphics[width=\linewidth]{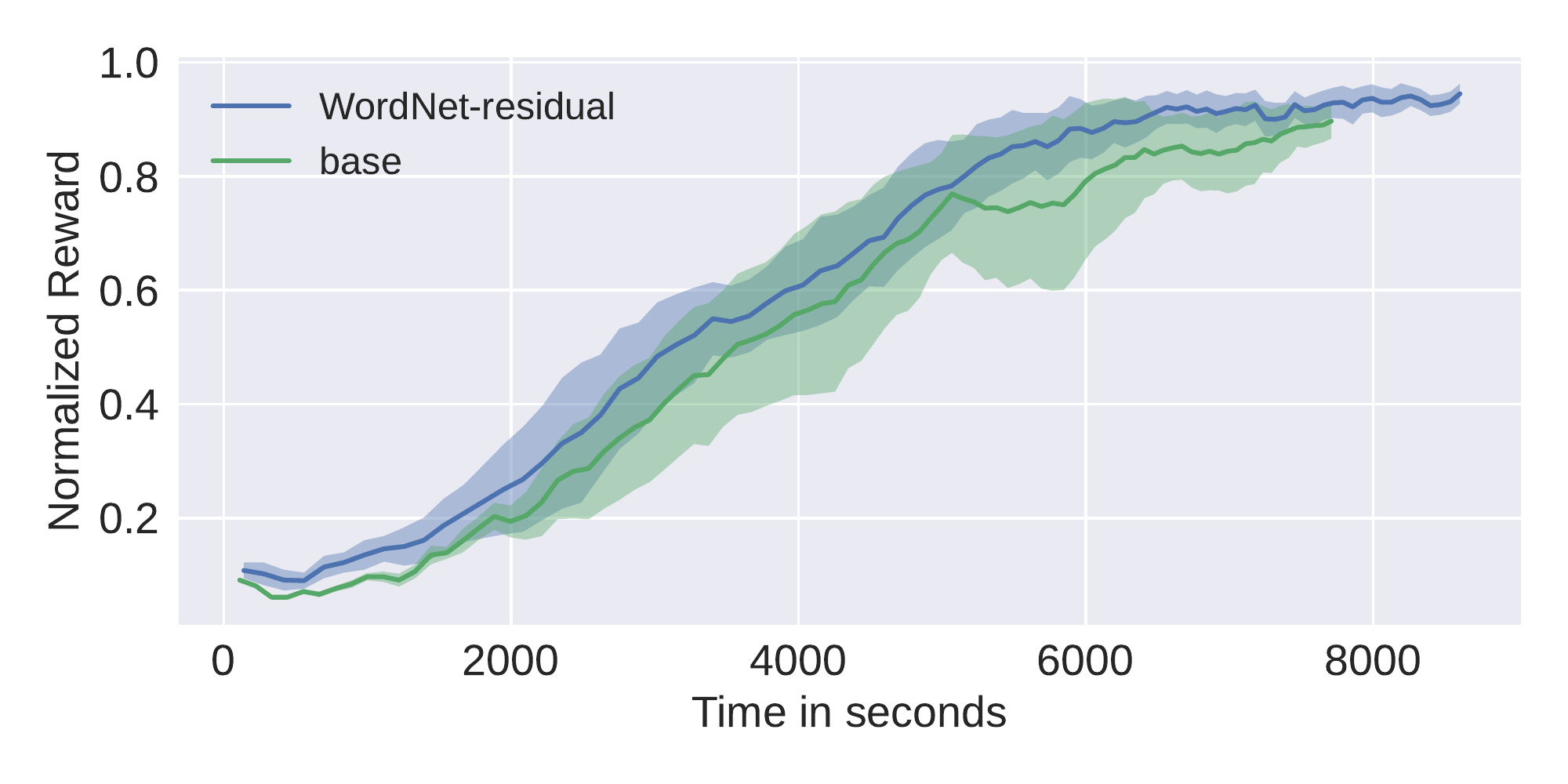} 
\caption{Normalised reward after time in seconds for the baseline and for the residual approach if abstraction is added from WordNet.}
\label{fig:time_plot}
\end{figure}

\begin{figure*}
\centering
\includegraphics[width=\linewidth]{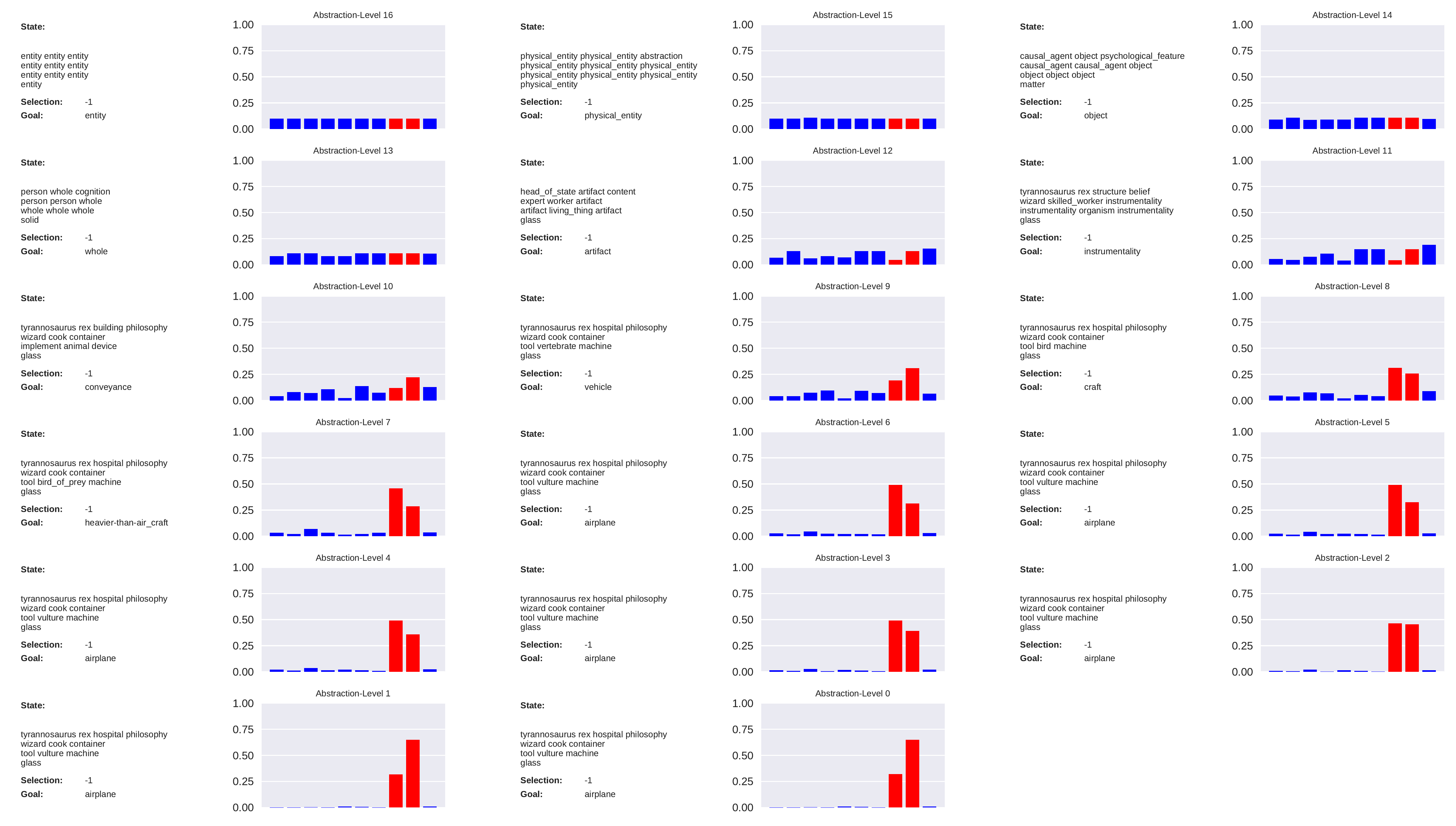} 
\caption{Visualisation of the policy trained via the residual approach, random embeddings and a class hierarchy extracted from WordNet. For each abstraction level the state representation is shown, which entity has been selected (-1 means no entity has been selected yet) and the abstracted goal entity. The probability with which each entity is picked is plotted and the red bars indicate the optimal actions.}
\label{fig:policy_visualisation}
\end{figure*}

Often an improved sample efficiency is achieved via an extension of an existing baseline, which introduces a computational overhead. If the method is tested in an environment where sampling is not the computational bottleneck it remains an open question whether the improvement in sample efficiency also translates into higher reward after a fixed time budget. To test this we measured the averaged time needed for one timestep over multiple seeds and fixed hardware and plotted the reward after time passed for the baseline and for the residual method with abstract states based on WordNet. All experiments were run on an a workstation with an Intel Xeon W2255 processor and a GeForce RTX 3080 graphics card. From Figure \ref{fig:time_plot} we can see that the baseline as well as the extension via abstraction achieve the same reward after a fixed time. Even if computation time is an issue the residual approach takes no more time than the baseline and its generalisation performance in terms of normalised reward is better.

\end{document}